\documentclass[sigplan]{acmart}
\usepackage{times}
\usepackage{blindtext}
\usepackage{tabularx}
\usepackage{graphicx} 
\usepackage{color}
\usepackage{xspace}
\usepackage{thumbpdf}
\usepackage{listings}
\usepackage{verbatim}
\usepackage{cleveref}
\usepackage{booktabs}
\usepackage{placeins}
\usepackage{balance}
\usepackage{subfig}
\usepackage{appendix}
\usepackage{amsmath}
\usepackage[small,compact]{titlesec}
\usepackage{url}
\hypersetup{breaklinks=true}
\usepackage[ruled,vlined]{algorithm2e}
\usepackage[subtle]{savetrees}
\usepackage[labelfont=bf,font=small]{caption}
\DeclareCaptionType{equationcaption}[][]

\newcommand{\oursys}{Catan\xspace}
\newcommand{\sr}{sense-react\xspace}
\newcommand{\Sr}{Sense-react\xspace}
\newcommand{\rt}{response time\xspace}

\newcommand{\gml}{GL\xspace}
\newcommand{\gm}{GM\xspace}
\newcommand{\gp}{GP\xspace}
\newcommand{\nc}{NC\xspace}
\newcommand{\lcm}{LM\xspace}
\newcommand{\lp}{LP\xspace}
\newcommand{\tw}{Timewarp\xspace}
\newcommand{\cam}{Camera\xspace}
\newcommand{\render}{Render\xspace}

\newcommand{\paragraphb}[1]{\noindent{\bf #1.}}
\newcommand{\paragraphi}[1]{\noindent\emph{#1.}}
\newcommand{\paragraphiq}[1]{\noindent\emph{#1}}

\setcopyright{none}
\renewcommand\footnotetextcopyrightpermission[1]{}
\AtBeginDocument{%
  \providecommand\BibTeX{{%
    \normalfont B\kern-0.5em{\scshape i\kern-0.25em b}\kern-0.8em\TeX}}}

\acmConference[]{}{}{}
\acmBooktitle{}
\acmPrice{}
\acmISBN{}

\begin{document}

%%
%% The "title" command has an optional parameter,
%% allowing the author to define a "short title" to be used in page headers.
\title{On-Device CPU Scheduling for Sense-React Systems}
\author{Aditi Partap}
\email{aditi712@stanford.edu}
\affiliation{%
  \institution{Stanford University}
%   \country{United States}
}
\author{Samuel Grayson}
\email{grayson5@illinois.edu}
\affiliation{%
  \institution{University of Illinois at Urbana-Champaign}
%   \country{United States}
}
\author{Muhammad Huzaifa}
\email{huzaifa2@illinois.edu}
\affiliation{%
  \institution{University of Illinois at Urbana-Champaign}
%   \country{United States}
}
\author{Sarita Adve}
\email{sadve@illinois.edu}
\affiliation{%
  \institution{University of Illinois at Urbana-Champaign}
%   \country{United States}
}
\author{Brighten Godfrey}
\email{pbg@illinois.edu}
\affiliation{%
  \institution{University of Illinois at Urbana-Champaign}
%   \country{United States}
}
\author{Saurabh Gupta}
\email{saurabhg@illinois.edu}
\affiliation{%
  \institution{University of Illinois at Urbana-Champaign}
%   \country{United States}
}
\author{Kris Hauser}
\email{kkhauser@illinois.edu}
\affiliation{%
  \institution{University of Illinois at Urbana-Champaign}
%   \country{United States}
}
\author{Radhika Mittal}
\email{radhikam@illinois.edu}
\affiliation{%
  \institution{University of Illinois at Urbana-Champaign}
%   \country{United States}
}

%%
%% The "author" command and its associated commands are used to define
%% the authors and their affiliations.
%% Of note is the shared affiliation of the first two authors, and the
%% "authornote" and "authornotemark" commands
%% used to denote shared contribution to the research.
%\author{\# 31}

%%
%% By default, the full list of authors will be used in the page
%% headers. Often, this list is too long, and will overlap
%% other information printed in the page headers. This command allows
%% the author to define a more concise list
%% of authors' names for this purpose.
%\renewcommand{\shortauthors}{\#XYZ, et al.}

%%
%% The abstract is a short summary of the work to be presented in the
%% article.

\begin{abstract}

 %\radhika{other suggestions for title, and for the app category (instead of sense-react)?}

\Sr systems (e.g. robotics and AR/VR) have to take highly responsive real-time actions, driven by complex decisions involving a pipeline of sensing, perception, planning, and reaction tasks. These tasks must be scheduled on resource-constrained devices such that the performance goals and the requirements of the application are met. This is a difficult scheduling problem that requires handling multiple scheduling dimensions, and variations in resource usage and availability. In practice, system designers manually tune parameters for their specific hardware and application, which results in poor generalization and increases the development burden. In this work, we highlight the emerging need for scheduling CPU resources at runtime in sense-react systems. We study three canonical applications (face tracking, robot navigation, and VR) to first understand the key scheduling requirements for such systems. Armed with this understanding, we develop a scheduling framework, \oursys, that dynamically schedules compute resources across different components of an app so as to meet the specified application requirements. Through experiments with a prototype implemented on a widely-used robotics framework (ROS) and an open-source AR/VR platform, we show the impact of system scheduling on meeting the performance goals for the three applications, how \oursys is able to achieve better application performance than hand-tuned configurations, and how it dynamically adapts to runtime variations. 

%This work takes a step towards \sr systems that are easier to develop and fulfill their tasks measurably better.

%Given constrained resources, this leads to a difficult scheduling problem. In practice, system designers manually tune parameters for their specific hardware and application, while real-time scheduling approaches assume static periodic schedules. Both approaches result in suboptimal application performance especially when both the environment and the hardware can change.

% our approach improves application performance, for example, 15x better performance for face tracking robot and 7x better collision avoidance for a navigation robot. 

%\radhika{abstract needs to be updated}

%\pbg{Is it OK to mention ILLIXR or is that somewhat violating anonymity?}

\end{abstract}

%%
%% The code below is generated by the tool at http://dl.acm.org/ccs.cfm.
%% Please copy and paste the code instead of the example below.
%%

\maketitle

\sloppy 
\section{Introduction}

%\radhika{Need to re-write to better align with the rest of the paper.}

\Sr systems
%such as robotics and AR/VR
are becoming ever more pervasive. A variety of robots now assist us with various mundane, high precision, or dangerous tasks~\cite{roomba, kiwi, amazonrobotics, surgrobot1, surgrobot2, surgrobot3, drone-fire, waymo, tesla}. AR/VR (Augmented/Virtual Reality) has affected the way we teach, practice medicine ~\cite{surgeon-xr, vr-neuro}, and play games, and is envisioned to be the next interface for compute~\cite{HuzaifaDesai2021, nvidia-xr}. These systems typically feature a pipeline of tasks that involve continually sensing the environment and processing the sensed inputs to generate a reaction. 
%sense-perceive-plan-react \sva{technically, there is no planning in AR/VR since the human user effectively does the planning? Or is the application rendering etc. analogous to planning? Have to think about this analogy} pipeline which 
Such a pipeline ranges from a single linear chain of components in the simplest applications to a multi-chain directed acyclic graph (DAG) in more complex ones. Most such systems run on devices with limited compute resources, (e.g. Intel NUC~\cite{locobot, terrasentia}, Raspberry Pi~\cite{turtlebot3-pi, turtlebot3}, Qualcomm Snapdragon XR2~\cite{qualcommVRPlatform}, etc.). 
% \radhika{is there a more generic name for such platforms?}~\sam{I think they are ``embedded systems,'' but I'm not sure about the mini PC.}. 
%\radhika{add some cites for HMD in AR/VR systems}.
% Run on edge systems : limited compute and memory resources. \todo{!!}
This work focuses on managing the compute (CPU) resources on such platforms. 
%In this work, we motivate the need for automated and dynamic resource management in such compute platforms, identify the key challenges, and develop a scheduling framework to dynamically manage the on-device CPU resources at runtime.

%Despite the rich literature in system scheduling, past work falls short of addressing the specific challenges in \sr systems. Most prior systems \radhika{add cites} assume a fixed  but in \sr systems, the environment changes continuously and is sampled at a configurable rate.

%\radhika{start with configurable rates -- Sam's comment.}
%\sam{I think we want to say: past work[citations galore] assumes inputs arrive at a rate determined by the environment, but in \sr systems, the environment changes continuously and is sampled at a rate determined by the system. In such a system, how fast should each component run, including the input nodes?}
%Unlike conventional systems, where the set of inputs 
\Sr systems allow the environmental inputs to be sampled and processed at a configurable rate (i.e. the input rate in \sr systems can be actively controlled, as opposed to being driven by external factors).
%(in contrast to conventional stream processing systems where input arrival rates depend on external factors \radhika{add cites}). 
CPU scheduling for a given \sr application (app) thus involves tackling two inter-related aspects --- how should the available compute resources be divided across different app components, and the rate at which each component should be executed (which includes the input sampling rate). 
%, and the order in which they run . 

The appropriate scheduling decision depends on the amount of available CPU resources, the compute usage of each component, and the app's performance requirements. 
% \aditi{the best scheduling decision also varies over time - often at multiple time scales, as there can be variations in compute usage and compute availability}
The scheduler must take into account dynamic variations in compute usage (due to 
%input-specific optimizations \pbg{that term isn't clear to me} triggered at various points, or 
% accumulation of more information about the environment, or a change in scene, etc.)
input-dependent computation times) and compute availability (e.g. due to battery constraints). 
The performance requirements differ along different components of a \sr app, and involve semantic trade-offs, where the tasks performed by some app components are more critical than others (e.g. in a navigating robot, components responsible for avoiding local collisions are more important than those responsible for global path planning). 
%The performance requirements along different components of an application are tied to semantics of the specific application, 
%which increases the difficulty of scheduling \radhika{fix this line}. 

As we show in \S\ref{sec:eval}, under-utilizing or overloading the system (e.g. by triggering a component too slowly or too fast), and mis-allocating resources (e.g. giving a greater share of resources to a less critical %\aditi{e.g. greater resources to planning the path than to avoiding collisions} 
component) impact the app's performance (that includes the ability of a robot to track a moving object or avoid collisions, or achieving the  desired motion-to-photon latency in an AR/VR system).
%\radhika{my todo: fix this paragraph}

Despite rich literature in system scheduling, past work falls short of addressing all of the specific challenges of \sr systems (e.g the ability and need to configure input rates, handle heterogeneous tasks with differing requirements, and tackle variability in compute usage at different time-scales). 
%They either optimize a subset of the dimensions listed above driven by a subset of performance metrics (e.g. just the triggering frequency ~\cite{davare2007period}, or the execution ordering ~\cite{yang-priorities, waters-priority, rosch-rt-e2e-lat}), or consider local (per-component) adaptations that do not take system-wide preferences into account \cite{aurora-load-control, borealis, seda, zeromq, rabbitmq, seda-followup, heron}, or rely on static configurations learnt offline that cannot adapt to dynamic variations  \cite{opentuner, automagic, bo4co, videostorm, auto-clustersz-hadoop} \radhika{revisit this paragraph -- do we need to soften it or is it accurate?}. \sva{I agree about revisiting the above para.} To the best of our knowledge, none of the prior work evaluate the impact of scheduling decisions on an applications' performance, particularly in the context of \sr systems. \sva{The last sentence seems too strong, depending on what is intended by "performance." For example, if performance is frame drop rate in a video teleconferencing application, then there are a large number of papers that do real-time scheduling and measure the frame drops, including our own prior work on GRACE.}
Thus compute resource management in such systems remains a challenge in which app developers are provided little help. While there are frameworks and systems that assist in app development (e.g. ROS~\cite{ros}, ILLIXR ~\cite{HuzaifaDesai2021,illixr-site}, etc), they leave scheduling decisions entirely up to the developers. Developers, therefore, manually fine-tune their systems to come up with static configurations which generalize poorly across scenarios (and over time) and increase the burden of app development. 
%Our goal is to ease the development and deployment of \sr apps, by providing a scheduling framework that can dynamically 

To ease the burden of app development and deployment, we build a scheduling framework, \oursys, which takes an app's semantic requirements as initial inputs from the developer, and dynamically schedules the app components at runtime as per the (varying) compute usage and availability. 
%The initial semantic inputs 
It is easier for the developer to specify the semantic inputs (that are based on domain expertise, and remain unchanged over time and across compute platforms), as opposed to directly configuring the scheduling knobs (that must be adapted based on compute usage and availability).  

%assists the app developer by determining the schedule across different app components. \oursys allows app developers to specify their semantic requirements (which remain unchanged over 

%allows an app developer to specify their semantic requirements, and  
%helps an app developer by dynamically scheduling compute resources at runtime across different components of a given \sr app, so to achieve the desired performance trade-offs and requirements. \radhika{my todo: fix -- something about semantic static inputs, and dynamic schedule}

We begin by studying three different applications (\S\ref{sec:apps-bg}) -- face tracking, robot navigation and exploration, and virtual reality. Through our case studies, we identify the key design considerations for a \sr system scheduler (\S\ref{sec:requirements}), and use the resulting insights to design \oursys (\S\ref{sec:design}). \oursys adopts a hierarchical approach, making its scheduling decision in two stages -- first determining the spatial allocation of CPU cores across app components, and then determining the temporal allocation of CPU slices and the rate of executing different app components.
%scheduler that appropriately allocates on-device CPU resources across different app components, running each component at the appropriate rate, as per the app's requirements and compute resource availability \radhika{my todo: fix this line}. 
We develop analytical models and heuristics to translate the scheduling decisions into light-weight constraint-optimization problems, that \oursys solves periodically at runtime to account for variability in compute usage and availability. \oursys can be plugged into existing frameworks for robotics and AR/VR  (e.g. ROS and ILLIXR respectively), and can be provided as an optional service to the apps using the framework (\S\ref{sec:impl}).

Through our evaluation on the three aforementioned case-studies (\S\ref{sec:eval}), we show (i) how scheduling decisions impact application performance, and how \oursys can (ii) effectively navigate the semantic trade-offs in performance goals as per the available compute resources, (iii) achieve better performance than the default (hand-tuned) configurations, and (iv) handle variability in compute usage at different timescales.

\section{Background}
\label{sec:background}

\paragraphb{Overview} We can represent a \sr application as a 
%Most \sr applications (particularly in the domain of robotics and AR/VR) can be characterized by 
directed acyclic graph (DAG), where each node (or vertex) represents a computation task and each edge represents the flow of data between tasks. The source nodes in the DAG (with no incoming edges) are comprised of various sensors, such as a camera, LiDAR, inertial measurement unit (IMU), etc., that continually capture environmental inputs. The sink nodes (with no outgoing edges) produce reactions, e.g. actuators in robotics and display in AR/VR. The in-between nodes are responsible for processing the input streams, e.g. by running detection and planning algorithms to determine appropriate reactions to changes in the environment. In most cases, every time a node runs, it uses the latest outputs generated by its predecessor nodes as its inputs (although some apps may have nodes that buffer multiple inputs and batch process them every time they run). DAGs for different applications vary in their complexity. We describe three examples in \S\ref{sec:apps-bg}.

%Each node in the DAG either runs at a certain configurable frequency or gets triggered upon the arrival of a new input from one of its predecessor nodes. 

%\radhika{check} 
%, that we use as our case-studies throughout the rest of the paper. 
%, and also discuss the task - specific goals for each of them.

% player might lose points if they cross the fault line.

% The main goal of the application is to provide a smooth visualization, which requires estimating the user's pose correctly [so that the rendered image is in the right pose], the accuracy of the displayed image w.r.t 

\paragraphb{Deployment Platforms} In many low cost robots, sensors and actuators are attached to a single on-board computer that runs all components in the DAG, often using CPU as the only compute resource~\cite{turtlebot, locobot, turtlebot3}. GPUs and other accelerators can be attached at the cost of higher expense and battery usage. 
%Note that this means that there would be contention among the different components for usage of the CPU. 
In larger robot systems, the nodes may be split across multiple on-board machines~\cite{fetch-doc, terrasentia}, or
some nodes may be offloaded to edge or cloud servers~\cite{cloudrobotics, cloudrobotics2, rapyuta, fog-robotics}.

While VR headsets have historically offloaded the computation to an attached server~\cite{htcvivepro}, the desired solution is to run AR/VR applications on stand-alone devices (e.g. Oculus Go~\cite{oculus-go} and Quest~\cite{oculus-quest}) equipped with an embedded system~\cite{qualcommVRPlatform, nvidiajetson, snapdragon835}. The embedded platforms provide on-board compute: CPUs, GPUs and accelerators such as DSPs.

In this work, we focus on systems using a single on-board computer, with CPU as the only \emph{contended} resource.~\footnote{
%Our robotics case studies use CPU as the only compute resource. 
Our VR case-study makes use of GPUs for frame display, but, unlike the CPU cores, we assume GPU resources are not contended.} This covers a broad range of real-world \sr systems. Extending our work to systems with multiple contended resources (CPU, GPU, memory, and network) is an interesting future direction.

\paragraphb{Development Frameworks}
It is common to use a software \emph{framework} (e.g.~\cite{orocos, lcm, ros, corba, yarp, nvidia-isaac, HuzaifaDesai2021, openxr}) for developing robotics and AR/VR applications.
%The modularity of such frameworks allow developers to easily re-use existing software packages, swapping specific nodes with newer ones, as needed.
Robot Operating System (ROS)~\cite{ros, ros-paper} is by far the most popular development framework for robotics, with more than 300K estimated users~\cite{ros-metrics}. 
It is widely used for developing research prototypes, with industry usage also growing rapidly~\cite{kiwi, fetch, clearpath, aws-robomaker, ros-cruise}.
It allows developers to program each component of an application individually, and provides communication APIs among those components.
% In ROS, each component can be implemented as either a node (i.e. a separate process) or a nodelet (thread within a process).
ILLIXR (Illinois Extended Reality testbed) is a complete end-to-end open source VR system and research testbed, which provides configurable VR system configurations with widely used system components and workflows~\cite{HuzaifaDesai2021,illixr-site}.
%, along with an extensible communication interface and runtime. 
ILLIXR provides an OpenXR~\cite{openxr} interface to the apps, which abstracts over the XR device, runtime, and OS.
%\radhika{fix this sentence}.
% All components are developed as plugins (or threads) within the same process.
%Both ROS and ILLIXR allow developers to program each component of an application individually, and provide communication APIs among those components.
% \sva{I would remove the above sentence -- it is confusing for illixr since these components are not what an app means; I think it is unnecessary for this paper. But its ok if you want to keep it. I think one source of confusion in this paper is what is that the meaning of an app is different for vr and robotics, but we can fix that later.}

%\sva{Again, fix. Need to say XR space has been proprietary (we dont know what the companies do for scheduling), illixr is the first open-source runtime etc.} 
The above (open)
%\footnote{Prior to~\cite{HuzaifaDesai2020}, XR platforms have been proprietary, and have not revealed their scheduling methods.} 
frameworks leave resource management decisions entirely up to the app developers. 
%\radhika{add/edit footnote to emphasize that we do not know what proprietary frameworks do.} 
Developers manually configure the rate at which different nodes are triggered. CPU scheduling across nodes is left up to the default OS policies, unless explicitly configured by the developer.~\footnote{We do not know what scheduling knobs and policies are used in closed-source / proprietary systems, which are heavily engineered nonetheless.} 
%We aim to design a scheduler for \sr applications that can be provided as a plug-in feature for such frameworks, and that can appropriately schedule the on-device CPU resources across different application components. 
%\radhika{rewrite last part to better explain what is hard about manual config., especially in context of where our scheduler helps.}

\paragraphb{Related Dataflow Systems} DAG abstraction is common across other systems, including real-time ~\cite{openvx-realtime, realtime-parallel-sched, dag-scheduling-rtc, yang-priorities,davare2007period,rosch-rt-e2e-lat,verucchi2020latency}, sensor nodes ~\cite{sensornode-pixie, sensornode-eon}, and distributed stream processing and dataflow systems  ~\cite{seda, borealis, heron, zeromq, rabbitmq, seda-followup, streamscope, aurora-load-control, sched-variable-load, aurora-load-control, load-shedding-data-streams, dfg-dryad, dfg-naiad}. \Sr systems are distinct from these. 

As we show in \S\ref{sec:requirements}, the computation (execution times) in \sr systems exhibits a high degree of variability over time --- an aspect that conventional real-time systems (based on fixed periodicity) do not handle. In comparison with traditional sensor nodes, \sr systems have higher processing complexity and variability, which emphasizes the importance of proper CPU scheduling. 

Distributed stream-processing systems focus on cluster-wide resource management to handle an incoming stream of queries. We instead focus on CPU scheduling within a single computer on board a robot or a VR device, that runs a single long-running application. Such a system differs in its scheduling knobs and requirements (detailed in \S\ref{sec:requirements}).
%, when compared to cluster-wide distributed dataflow systems. 
For instance, in \sr systems, the scheduler has greater control over the app, and can actively control the input load.
%by changing the rate at which the source nodes sense the environment. 

%Finally, the long-running nature of the applications allows for periodic monitoring of resource usage and availability, and consequent updates in the scheduling decisions.

To summarize, \sr systems are more dynamic than traditional real-time systems, more complex than sensor nodes, and more controlled than distributed dataflow systems. They thus open an interesting design space for system scheduling. %\radhika{revisit this sentence}.

\section{Our Case Studies}
\label{sec:apps-bg}

%\radhika{motivate why we use these apps as case-studies. maybe move expt/app setup from eval to this section. some of the text here is difficult to understand, so rewrite accordingly.}

\begin{figure}[t]
\centering
\includegraphics[width=0.35\textwidth,trim={0 0 0 0},clip]{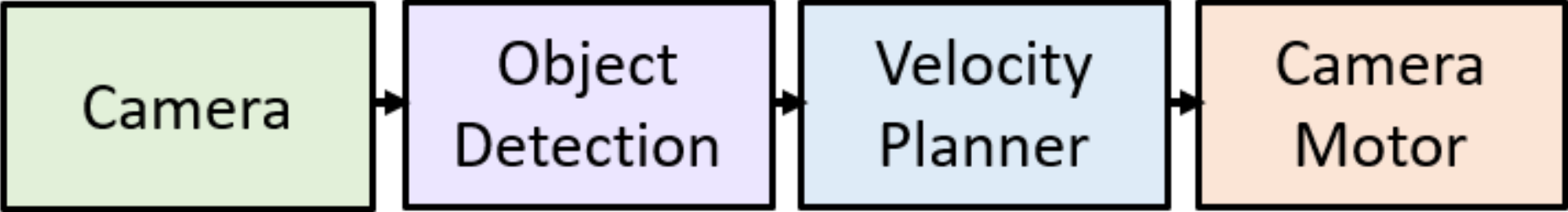}
\vspace{-10pt}
\caption{DAG representing the face tracking robot app. The colors green, purple, blue and orange represent sensor nodes, perception nodes, planning nodes and actuators respectively}

\label{fig:objtrack-dag}
\end{figure}

We use the three representative applications, spanning robotics and AR/VR, as our case studies to inform and evaluate our scheduler design. We briefly describe these applications below. 

%In this section, we provide an overview of these applications. 

\subsection{Face Tracking Robot}
\label{sec:facetrack-bg}

As our first case study, we consider a face-tracking robot implemented using ROS~\cite{goebel-2013, ros-face-tracking}. It involves a rotating camera tasked with tracking a moving object (face). 
%It is a simple app that helps highlight the importance of basic scheduling decisions made even in the absence of complex semantic trade-offs and high degree of variability.
Figure~\ref{fig:objtrack-dag} represents the app DAG, which is a simple linear pipeline of the following nodes: (i) a camera to capture and preprocess images, (ii) an object detection node to detect the face in the image, (iii) a planning node to compute the velocity at which the camera must rotate to track the face, and (iv) the camera rotation motor. 

 %\radhika{revisit this sentence} 

% \radhika{add camera motor to the figure?}

\paragraphb{Performance Goals} The goal of this robot is to always keep the object in its view, i.e. to not lose track of the object.

% In more complex applications, the linear pipeline evolves into a multi-chain DAG.

\begin{figure}[t]
\centering
\includegraphics[width=0.45\textwidth]{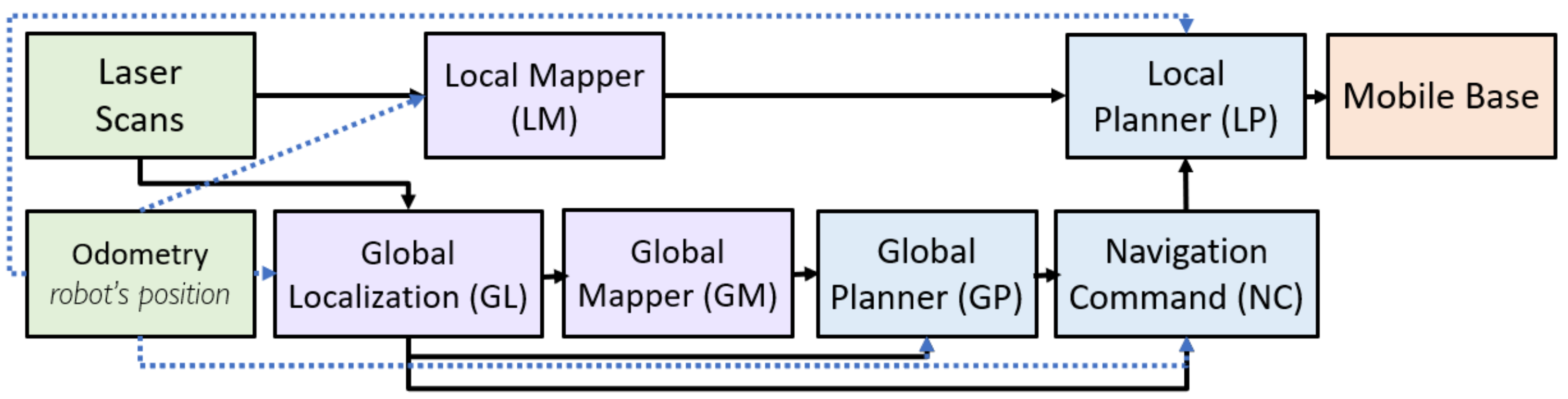}
\vspace{-10pt}
\caption{DAG representing the robot navigation application.}
\vspace{-10pt}
\label{fig:nav2d-dag}
\end{figure}

\subsection{Robot 2D Navigation}
\label{sec:navigation-bg}

For our second case study, we consider a robot navigation app implemented using ROS~\cite{nav2d}, where the robot is tasked with exploring and mapping an unknown area. This is a basic feature that most autonomous mobile robots need to support \cite{prbook, yamauchi1997frontier}. This is a more complex app, that involves semantic trade-offs, and higher variations in compute usage over time. 

Figure ~\ref{fig:nav2d-dag} represents the application's DAG. It uses two sensors -- a laser scanner (to capture the environment seen by the robot) and an odometer (which reports speed and location information to all other nodes). The global localization, mapping and planning nodes (\gml, \gm, and \gp) are responsible for planning the robot's trajectory based on its accumulated knowledge about the area. 
% It plans the trajectory 
The trajectory is planned so as to move towards unknown areas for exploration. The local mapping and planning nodes (\lcm and \lp) are responsible for ensuring that the robot avoids collision with obstacles in its immediate vicinity when following the global trajectory. We list the specific function of each processing node below. More details about relevant algorithms can be found in \cite{prbook, aibook}. 
%\radhika{check if GML can be replaced with GL?}  

\paragraphi{(i)} The local mapper (\lcm) uses the laser scans and odometry to update its knowledge about the robot's immediate vicinity. 

\paragraphi{(ii)} The global  localization node (\gml) performs two tasks: (a) for every scan that it receives, it uses particle filtering (Chapter 8 in \cite{prbook}) to produce an estimated correction for the robot's location (which accounts for potential drifts in odometry readings), and (b) it then filters the scans, discarding the ones that carry little new information about the environment. Unlike other nodes in the app that always use the latest inputs available from their predecessor nodes, \gml buffers the received scans, and batch processes them every time it runs to ensure that all relevant information about the environment is captured. 
%for effective exploration and path planning.

\paragraphiq{(iii)} The global mapper (\gm) maintains a global map of all the areas that the robot has explored so far. It uses occupancy grid mapping (Chapter 9 in \cite{prbook}) to update the map based on newly filtered scans produced by \gml. 
% corrects the robot's location obtained from the odometer, and pre-processes the laser scans 
 
\paragraphiq{(iv)} The global planner (GP) computes the global trajectory of the robot based on \gm's global map and the robot's position (derived from the latest correction from \gml and the corresponding odometry reading) using graph search (Chapter 3 in \cite{aibook}).

\paragraphiq{(v)} Navigation Command (\nc) uses the robot's position, along with the current trajectory (i.e. \gp's output) to decide the direction in which the robot should move.
% which is translated into a direction by the navigation command node (NC). 

\paragraphiq{(vi)} The local planner (\lp) uses the local cost map (that contains information about nearby obstacles) and the odometry information, along with the navigation command from \nc, to output the robot's velocity to the actuator (the mobile base).
% \radhika{I edited this. Check if this correct}

%\sg{if space, add something about the algorithm used by the above nodes, noting that others are possible.}

\paragraphb{Performance Goals} The foremost goal is to avoid colliding with obstacles. The secondary goal is to ensure that trajectories are planned using the most up-to-date information about the robot's location and the map, otherwise planned trajectories will be infeasible, unsafe, and inefficient. Finally, the robot should be able to explore the area at a sufficiently high rate.

\begin{figure}[t]
\centering
\includegraphics[width=0.35\textwidth]{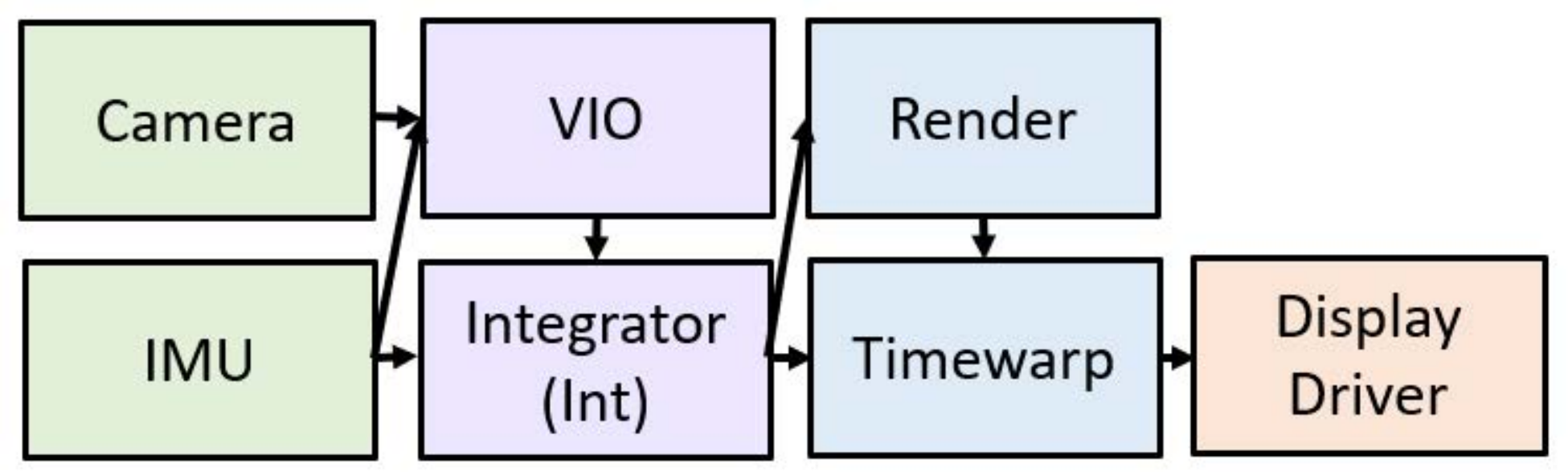}
\vspace{-10pt}
\caption{DAG representing the VR application in ILLIXR~\cite{HuzaifaDesai2021,illixr-site}}
\vspace{-14pt}
\label{fig:illixr-dag}
\end{figure}

\subsection{Virtual Reality} 
\label{sec:vr-bg}

Our final case study considers a different category of \sr systems -- virtual reality (VR) implemented in ILLIXR~\cite{HuzaifaDesai2021, illixr-site}. The VR system's task is to provide a smooth visualization of the virtual world, based on the user's head movement in the real world. 

Figure ~\ref{fig:illixr-dag} represents the various nodes in a VR system. It uses two sensors: (a) an IMU (inertial measurement unit), which measures acceleration and angular velocity of the user's head, and (b) a \cam.
%, which captures images of the surrounding environment. 
The Visual-Inertial Odometry node (VIO) uses a localization (SLAM) algorithm to fuse information from the camera and IMU, and produce an estimate of the user's pose (location and orientation).
%\footnote{It is similar to the \gml component from the robot navigation application.}
It generates high accuracy poses but is computationally too expensive to run at a high frequency.  
%but only does so at the frequency of the camera, which is typically not high enough for VR applications. 
To remedy this, the integrator (Int) integrates new samples from the IMU to estimate the change in the pose since the last pose from the VIO. This produces higher frequency, but less accurate, estimates of the user's pose.

The Render node uses this pose to generate an image (aka a `frame') of the virtual world from the current perspective of the user.
% However, the user's pose has typically changed by the time the rendering process finishes.
However, the latency of the rendering process adds a lag between the user's movement and the generation and display of the corresponding frame, called the motion-to-photon latency (MTP)~\cite{Waveren2016,daqriATW}.
To minimize this latency, the Timewarp node takes the latest rendered frame from Render and the latest pose estimate from Int, and reprojects the frame to be from the latest perspective of the user  
(reprojecting is much faster than rendering a new frame).
%so it can be scheduled just before the display outputs a frame. 
Timewarp submits these reprojected frames to the GPU driver which displays them on a head-mounted display at the refresh frequency, based on the display hardware.
%Furthermore, if Render is unable to finish rendering on time (a dropped frame), Timewarp uses the last available frame for reprojection. Thus, Timewarp is a critical component in VR as it ensures that frames are always presented from the correct perspective of the user and with minimal latency.

The reprojection in the ILLIXR version we use   accounts for the user's rotational motion, but not translational motion
% (from the time corresponding to the pose used by render).
%
%, \fixme{sam: maybe high-level reason, o.w. could remove} since different objects in the virtual world parallax by different amounts according to their radial distance
% which has to go through render.
Therefore, we report two MTP metrics - rotational MTP and translational MTP, representing the delay in capturing the user's rotational and translational motion respectively.\footnote{It is difficult to capture the latency of the actual display; we therefore do not include that in our MTP, similar to~\cite{HuzaifaDesai2021}. With current display technologies, this latency is roughly constant and affects all scheduling mechanisms studied here similarly.
%\sva{Check this footnote. Should we mention that timewarp also does pose prediction for the end of display pose, or is that too much detail?}
%\sam{I think the appropriate amount of detail is already reflected in that timewarp takes input from imu integrator. What it does with that input )pose prediction) is less relevant for scheduling purposes.}
} 

\paragraphb{Performance Goals}
To provide a smooth virtual experience, the primary goal of the scheduler is to ensure that the displayed frame 
closely tracks the user's motion with low delay and the best available pose.

\vspace{10pt}
Our experiment setup for each of the above case studies uses the apps as implemented in their respective frameworks (ROS for \S\ref{sec:facetrack-bg} and \S\ref{sec:navigation-bg}, and ILLIXR for \S\ref{sec:vr-bg}). In order to do controlled (repeatable) experiments, we use standard physics simulators ~\cite{gazebo,Stage} to simulate the sensing and actuation for the two robotics apps (as is standard practice in robotics ~\cite{robotics-simuln}) -- the rest of the processing nodes are executed as in a real system. 
%, thus doing ``real'' processing on simulated inputs.
% \footnote{This is standard practice in robotics ~\cite{robotics-simuln}}
% and also in VR~\cite{vr-simuln1, vr-simuln2}.
% For VR, we use camera and IMU inputs provided by  standard datasets (for repeatability), fed into the "real" ILLIXR VR system (which is agnostic to the source of these inputs).
For VR, ILLIXR supports using offline generated datasets as well as real-time camera and IMU data. For repeatability, we choose to use widely used sensor datasets~\cite{BurriNikolic2016}\footnote{We use the sensor dataset to emulate the possible trajectory of a person wearing the VR device. We used the aerial dataset since there is no available dataset for human head movement.}; the rest of the (real) ILLIXR system is agnostic to this choice.
% \sva{check my changes, I split robotics and VR, integrated the robotics footnote, and removed the vr simulation cites. We have a REAL VR system -- those vr cites are simulations that undermine our work.} \aditi{these look good}
% REVISIT THIS:
% \sva{Since we are using offline datasets, what is the work that camera and imu nodes are doing as reported in our experiments? Should we clarify that?}
% \sam{Those components are loading in offline data, some of which would not be present in live data (e.g. reading from disk), some of which would be (e.g. froamt conversion).} \aditi{lets not go into this due to space constraint...}\sam{That's what I think too. Not sure Sarita would be happy?} \aditi{we'll revisit this after one full pass and updating the submission?}
% for VR. We therefore do ``real'' processing on simulated/offline inputs.
% \aditi{we are going beyond 12pages right now (after removing the macro)...}
We detail the experiment setup for each case study in \S\ref{sec:eval}. %\radhika{rewrite}
\section{Design Considerations for CPU Scheduler}
\label{sec:requirements}

%\radhika{This section is very dense, and hard to follow. Re-write such that readers will be able to better appreciate the high-level points.}

%\radhika{better section title needed}

%We use our case studies to identify the minimal set of requirements that an on-device CPU scheduler for \sr systems must satisfy.~\footnote{These are minimal in the sense that they were derived from the three applications we studied. Analyzing more applications may reveal other requirements. Nonetheless, we believe that our study combining different components across the three representative apps provides a broad coverage, and a solid starting point for the scheduler design.} \radhika{update}

We use the above case studies to inform our scheduler design. In this section we list some of the key design considerations. 

%\subsection{DAG Terminology} 
%\label{sec:terms}

\vspace{5pt}
\paragraphb{Terminology} First, we will clarify terminology. As mentioned in \S\ref{sec:background}, we use the term \emph{node} to refer to each vertex of the application DAG. 
We use the term \emph{chain} to denote a unique path from a source node (e.g., a sensor) to a sink node (e.g., the actuator or display) along the DAG. A node may belong to multiple chains. For example, the face-tracking app (Figure~\ref{fig:objtrack-dag}) comprises of a single chain from the Camera to the Motor. On the other hand, the navigation app (Figure~\ref{fig:nav2d-dag}) consists of multiple chains from scan and odometry to the mobile base, that pass through different sets of intermediate nodes (e.g. \{scans $\rightarrow$ \lcm $\rightarrow$ \lp $\rightarrow$ base\}, \{scans $\rightarrow$ \gml $\rightarrow$ \nc $\rightarrow$ \lp $\rightarrow$ base\}, \{scans $\rightarrow$ \gml $\rightarrow$ \gp $\rightarrow$ \nc $\rightarrow$ \lp $\rightarrow$ base\}, etc). Likewise, the VR DAG (Figure~\ref{fig:illixr-dag}) consists of six different chains from Camera and IMU to Display, passing through different intermediate nodes.

\subsection{Scheduling dimensions}
\label{sec:dimensions} 

%In order to optimize app performance the CPU scheduler must co-optimize different scheduling dimensions.  

%We first discuss what CPU scheduling in an \sr system entails. 
In order to optimize app performance, the CPU scheduler must co-optimize the following key dimensions. 

\paragraphiq{(i) Spatial Core Allocation.} Given a multi-core platform, the scheduler must determine how the CPU cores are divided across different components in a \sr app. %\sam{Perhaps call this Spatial Core Allocation, so it is complementary to Temporal Core Allocation.}

%We should also account for heterogeneity among cores, and hyper threading within each core \radhika{Aditi -- does our scheduler handle the latter -- if not, we should cut} \aditi{[not modelling right now, but can be handled easily i think]}.

\paragraphiq{(ii) Temporal CPU Allocation.} For a core to which multiple components are assigned, the scheduler must decide how many CPU slices must be allocated to each of them, and how often.
%(which follows from determining its execution rate), and \emph{when}. 

% \paragraphiq{(iii) Degree of parallelism.} Certain nodes in the application DAG may support multi-threading. 
% %For example, the \gm processing in the navigation DAG can be parallelized. 
% The scheduler must decide the number of threads that such a node is allowed to use. \sam{Is this different from spatial core allocation?}

\paragraphiq{(iii) Execution Rate.} For each component, the scheduler must decide the rate at which it is triggered (or executed). This follows from determining the CPU allocation for that component.

We use the term `component' to loosely refer to the granularity at which the scheduler makes its decisions -- we discuss this more precisely in \S\ref{sec:granularity}.

% \radhika{Aditi, based on Sam's comments, I removed ``degree of parallelism'' from here. In a way, the first point on spatial allocation covers it. We can talk about it directly in S5 -- this again helps in de-emphasizing it. Comment if you think we should mention it here.} 

%,  i.e. use inputs from upstream nodes to produce a new output.

%Each of the above decisions impact the system metrics discussed in \S\ref{sec:metrics}. Allocating more cores to a chain can increase its throughput (through better pipelining) and/or lower its processing latency (through higher per-node parallelism). However, this may come at the cost of fewer cores being allocated to other components, and must therefore take into account their relative weights and constraints (\S\ref{sec:preferences}). The execution rates should be appropriately computed so as to avoid overloading the allocated resources, meet the required constraints (e.g. the throughput upper bounds), and achieve the sweetest spot in the semantic trade-off space when sharing a CPU core with other components. The temporal ordering of components on a CPU core impact the latency of the associated chains.      

\subsection{Scheduling granularity}
\label{sec:granularity}

We use the term \emph{subchain} to refer to a series of DAG nodes within a chain that run at the same rate in an event-driven manner (with the output from one node triggering the next). Two nodes in a chain would belong to different subchains if there is value to running them at different rates (e.g. if they belong to a different set of chains). Note that each node may be in many chains, but can only belong to one subchain.
%The overall throughput (and consequently the response time) of a chain is bottlenecked by its slowest subchain.

For example, in the navigation DAG (Figure~\ref{fig:nav2d-dag}), it makes semantic sense for \lp to output a new velocity only upon every new input from \lcm that carries updated knowledge of the robot's immediate vicinity (and can typically arrive at a faster rate than global navigation commands from \nc). Therefore, \lp and \lcm can belong to the same subchain. On the other hand, it is useful to run \gp at a higher rate than \gm (which is computationally more expensive), allowing the global trajectory to be updated based on updated position estimates. 
%Likewise, it is useful for \nc and \gml to run at higher rates than \gm and \gp. Therefore, \gml, \gm, \gp, and \nc may each comprise of different subchains.  

%to run \lcm and \lp at the same rate, and trigger \lp only upon the arrival of a new output from \lcm (implying that the robot outputs a new velocity every time it  .
%On the other hand, 

Since all nodes within a subchain must run at the same rate, a subchain forms the natural granularity at which we can make the scheduling decisions listed in \S\ref{sec:dimensions}. The scheduler needs to actively trigger only the first node in each subchain, and each of the remaining nodes in the subchain can be event-triggered when the preceding node produces a new output.~\footnote{This has lower latency than asynchronous triggering of consecutive nodes.}

%This has a number of advantages: (i) Since different subchains run at different rates, \radhika{fill in after discussion}  (ii) Reasoning about subchains is more tractable than reasoning about individual nodes [\aditi{unsure why}].  (iii) It enables the scheduler to schedule the nodes within a subchain in an event triggered manner, which results in lower latency than triggering the nodes asynchronously using independent timers \radhika{more like enables the scheduler to get "out of the way". to discuss}.   

%\oursys makes scheduling decisions at the granularity of a \emph{subchain}. Each subchain is a series of nodes that run at the same rate in an event-driven manner (with the output from one node triggering the next). Two nodes in a chain would belong to different subchains if there is value to running them at different rates. \aditi{For example, in the navigation DAG, putting LC, LP in the same subchain helps avoid extra latency of LC's outputs waiting at LP. Also, since compute time of GP can be much higher than that of NC, it makes sense to keep them as separate subchains, so as to not restrict to run at GP's low frequency (since a node cannot run at a frequency more than the reciprocal of its compute time).} \oursys takes the subchains in the application DAG as an input. It can be extended in future to automatically determine how an application DAG can be split into subchains. 

\subsection{Performance metrics}
\label{sec:metrics} 

%\Sr apps have a variety of different performance objectives which are impacted by the above scheduling decisions. 
The scheduling decisions (\S\ref{sec:dimensions}) impact the performance goals listed in \S\ref{sec:apps-bg}. However, it is difficult for the scheduler to directly reason about app-level performance metrics (e.g. ability to avoid collisions or track a moving object). 
%We instead translate app-level metrics into generalized low-level system metrics, that the scheduler can directly reason about and optimize. 
%By studying the correlation between app performance and low-level system metrics, we identify reveals that the scheduler must support different metrics.
We instead express different app performance metrics using generalized DAG-level metrics that the scheduler can directly reason about and optimize. Our first metric corresponds to how quickly and frequently new inputs are processed along a given \emph{chain} in the DAG, while the second corresponds to the processing rate of a given \emph{node}. 
%The face tracking (Figures \ref{fig:objtrack-dag}), navigation (\ref{fig:nav2d-dag}) and VR (\ref{fig:illixr-dag}) applications in our case-studies have 1, 4 and 6 chains respectively \radhika{this might need to be fixed based on updated figure}. 
We describe these metrics below: 
%\radhika{revisit if we need to add chain latency as another metric. Say `two categories of metrics. First category captures ..., and  the second...'. Then the paragraph headings will be Chain-level metrics and Node-level metrics....}

%\paragraphiq{(ii) Chain Throughput} is the rate at which inputs from the source node are processed along a chain to generate new outputs at the sink. 
%It is measured as the reciprocal of the time gap between two consecutive outputs ($o_k$ and $o_{k-1}$) at the sink of the chain that use 

%\paragraphiq{(iii) Chain Latency} is the time taken for an input $i_k$ at the source to be processed along the chain and generate an output $o_k$ at the sink.

\paragraphb{Chain Response Time} It is the worst-case time from the moment a change occurs in the environment, to the moment a reaction is produced at the sink of the chain.
%\footnote{This metric shares similarity with ``age of information'', a data freshness metric defined in the context of network communication~\cite{aoi}.} 
We define response time for consecutive pairs of inputs ($i_{k-1}$ and $i_{k}$) that are fully processed by the chain to produce new outputs at the sink ($o_{k-1}$ and  $o_{k}$ respectively).~\footnote{Note that there may be other inputs captured by the chain source between $i_{k-1}$ and $i_{k}$ that get dropped (or overwritten by newer inputs) along the chain and do not influence the output at the sink. Likewise, there may be other outputs between $o_{k-1}$ and $o_{k}$ that use newer inputs along other chains incident at the sink, but are based on input $i_{k-1}$ from the given chain's source.}
%In the best-case, an environmental change would occur right before the sensor captures an input $i_{k}$ that gets processed by the chain to generate a reaction $o_{k}$ at the sink. 
In the worst case, an environmental change occurs immediately after the source of the chain captures a previous input $i_{k-1}$ --- the system will not react to the change until the next input $i_{k}$ is captured and processed by the chain to produce the output $o_{k}$. Chain response time is, therefore, the time difference between when the sink produces a new output $o_{k}$, and when the source captures the \emph{previous} input $i_{k-1}$. Intuitively, minimizing response time ($o_{k} - i_{k-1}$) implies simultaneously minimizing the latency of processing an input along a chain ($o_{k-1} - i_{k-1}$) and maximizing the chain throughput, i.e. minimizing the period ($o_{k} - o_{k-1}$) between two consecutive outputs.
%Response time, as a metric, is most relevant in chains where each node uses the latest input from the previous node (buffering inputs generally increases the chain latency and response time). 

%For the face tracking application (\S\ref{sec:facetrack-bg}) involving a single chain, 
%As highlighted in \S\ref{sec:eval}, 
The robot's ability to track the moving object for the face tracking app is correlated with the chain's response time (\S\ref{sec:eval}). Likewise, a navigating robot's ability to avoid collisions is correlated with the response time along the chain `scans $\rightarrow$ \lcm $\rightarrow$ \lp $\rightarrow$ base', which captures how quickly the robot can react to dynamic obstacles in its immediate vicinity. Minimizing rotational MTP in VR directly translates to minimizing response time along the `IMU $\rightarrow$ Int $\rightarrow$ Timewarp $\rightarrow$ Display' chain. 
%\radhika{revisit} 

\paragraphb{Node Throughput} It is the rate at which a given node processes its inputs to produce new outputs.
%, which In order to effectively track a moving object, it is important both to capture the environmental inputs (images) and produce reactions (motor commands) at a high enough rate, while ensuring minimal delay (latency) between when an input is captured and the corresponding reaction is produced. This makes response time the key metric that the scheduler must optimize. 
%For the navigation application (\S\ref{sec:navigation-bg}), we similarly find that the robot's ability to avoid collisions is correlated with the response time along the \emph{local chains} (`scans $\rightarrow$ LC $\rightarrow$ LP $\rightarrow$ base' and `odometry $\rightarrow$ LC $\rightarrow$ LP $\rightarrow$ base'), which captures how quickly the robot can react to dynamic obstacles in its immediate vicinity. On the other hand, 
%The \gml node in the navigation app (\S\ref{sec:navigation-bg}) feeds into multiple other nodes (and therefore belongs to multiple chains). 
Ensuring that the latest position estimate is available for generating the robot's trajectory requires high throughput for the \gml node (that feeds into multiple other nodes) in the  navigation app.

\subsection{Semantic preferences and constraints}
\label{sec:preferences} 

%The scheduler should provide a general interface to capture different semantic requirements.

\paragraphb{Preferences across different metrics} Certain performance metrics are semantically more important than others. For example, in order to avoid collisions during robot navigation, minimizing response times along the local chains must be given highest priority. Next in line is ensuring high \gml throughput (to avoid faulty navigation commands derived from stale position information), followed by a sufficiently high throughput for the remaining nodes to minimize the exploration time. Similarly, as detailed in \S\ref{sec:eval}, the VR system also involves semantic preferences across the response times for different chains.

%Likewise, for the VR application, minimizing latency \radhika{response time} along the `IMU $\rightarrow$ Int $\rightarrow$ Timewarp $\rightarrow$ Display' chain (corresponding to rotational motion-to-photon latency) is most important, followed by minimizing latency \radhika{response time} along `IMU $\rightarrow$ Int $\rightarrow$ Render $\rightarrow$ Timewarp $\rightarrow$ Display' chain (which influences translational \aditi{update} motion-to-photon latency), with VIO's throughput \radhika{or chain response time -- what do we report?} (that affects pose estimation accuracy) having the lowest priority. 

%The scheduler should provide an interface to app developers to assign different weights to each metric, so as to achieve the desired semantic trade-offs.
%without starving the lower priority components under limited resource availability. 

\paragraphb{Constraints on different metrics} 
%An app may further require the schedule to adhere to constraints on certain metrics. 
An app may require an upper bound on a chain's response time or a lower bound on a node's throughput, based on semantic requirements. In addition, a node's throughput can be upper bounded by the hardware limits of the sensors and actuators/display. 

\begin{figure}
\centering
  \setlength{\tabcolsep}{1pt}
  \renewcommand{\arraystretch}{0.9}
  \resizebox{\linewidth}{!}{
   \begin{tabular}{cc}
    % \includegraphics[width=0.2\textwidth]{cs_thesis_latex_template-2020/figures/GM_run1.pdf}
    % &
    \includegraphics[width=0.25\textwidth]{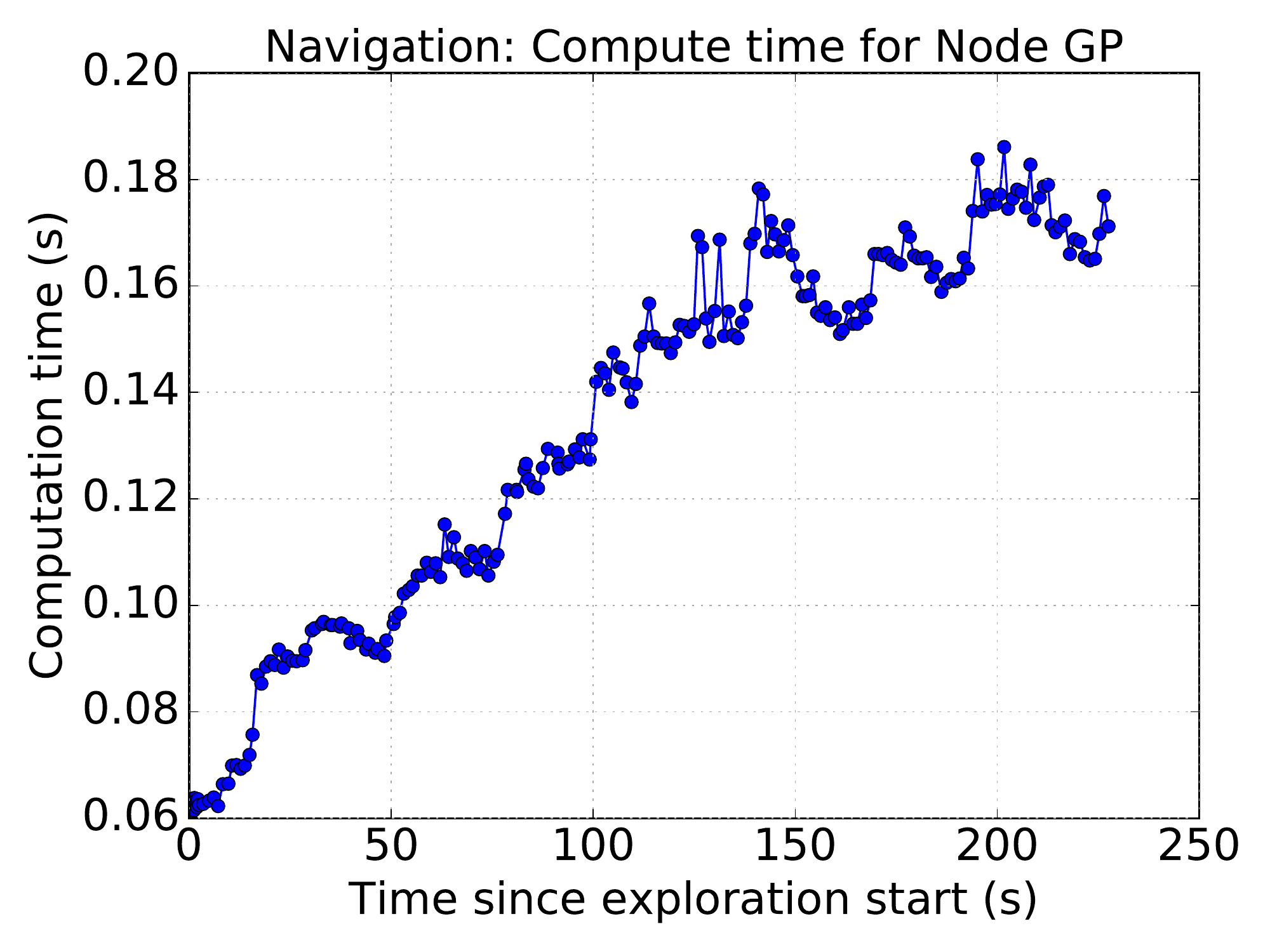}
    &
     \includegraphics[width=0.25\textwidth]{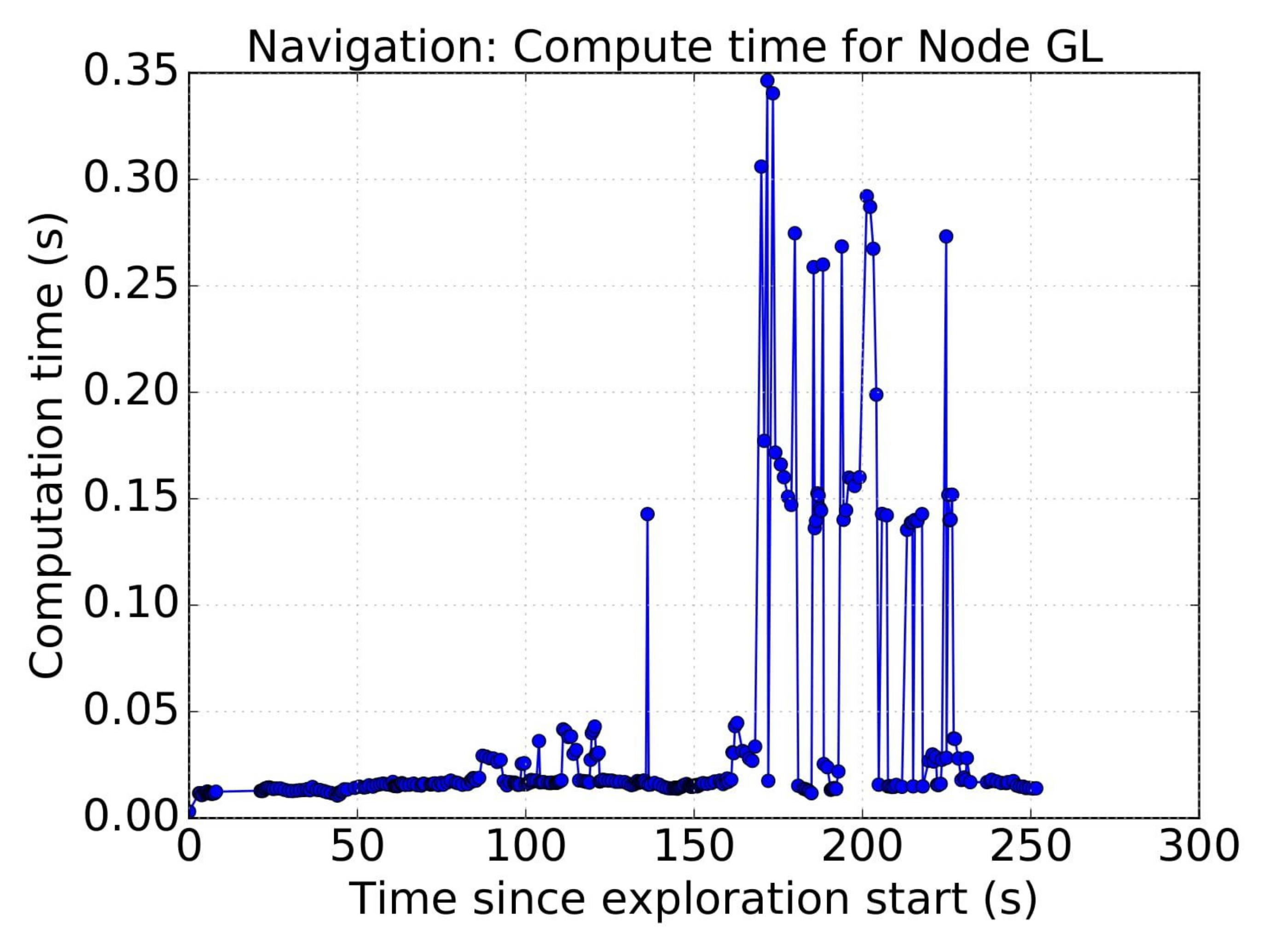}

    \end{tabular}}
    \vspace{-10pt}
    \caption{Variation in the CPU usage over time of two components in the navigation application: GP (left) and GL (right).}
    \vspace{-10pt}
    \label{fig:nav-compute-time-gp-gml}
\end{figure}

\subsection{Variations in Compute Usage and Availability}
\label{sec:variations}

%Dynamicity at different level -- over time and across scenarios.

%The scheduling decisions above would depend on the amount of CPU that each node consumes every time processes a new input. We call this the computation time of a node. 

%The scheduler should be able to account for variability in the 

%The node compute usage and availability can exhibit variability due to multiple reasons. 

%We observe that an \sr application typically exhibits variability in the amount of compute required by node to process its inputs every time it runs (we measure this as the node computation time). Further, the available amount of compute resources available may vary across scenarios as well. We briefly discuss the reasons for these variations.   

\paragraphb{Variation in compute usage over time}
We use node computation time to capture the amount of CPU consumed by a node every time it runs.  
 %measure the amount of compute required by node to process its inputs every time it runs as the node's computation time.
We observe that the computation time of some nodes may increase over time. For example, as the navigating robot covers more area, the size of the global map increases, which increases the time taken by (i) \gm to update and generate the global map, and  
%increases. Moreover, as the size of the global map increases, the time taken by 
(ii) GP to plan the robot's trajectory based on the map (as shown in Figure ~\ref{fig:nav-compute-time-gp-gml} (left)). 
%This is a common behavior across many robotics applications (e.g. the cost for checking collisions while planning motions for a manipulator grows with the number of convex obstacles \cite{mpbook}).  

\paragraphb{Input-dependent variability} We observe that the \gml node for robot navigation has a bimodal computation time --- for each laser scan, it either does a quick check and discards it if there is no new information, or processes the scan to update the pose correction.
%More generally, any node in a \sr system that performs semantic filtering (e.g. cascaded models~\cite{li2018low, haar}) would encounter such bimodality. 
%The scheduler must take such bimodal computation times into account when making its scheduling decisions.
%\radhika{how are we accounting for it? Make sure Sec 4 talk about it.}
For the frames that were fully processed by \gml, we additionally observe occasional spikes in computation times (as shown in Figure~\ref{fig:nav-compute-time-gp-gml} (right)). A closer analysis revealed that these spikes arise from a semantic optimization called ``loop closure''~\cite{slam-loopclos, prbook}, where on re-visiting a location, the robot runs an expensive non-linear optimization to jointly update the map and all robot poses. 

The VIO node in the VR app also shows high variability in compute usage, depending both on the complexity (feature density) of the user's surroundings and the user's motion.
%\radhika{Do you have a graph for VIO's computation time that we can put here?}

\paragraphb{Variation in available compute resources} The amount of resources available to the application would vary across different hardware platforms. Even for a given platform, the amount of available compute resources may change over the duration of the application run ---
%They might vary over the  but could also vary over time. 
e.g. as a robot's battery starts draining over time, the number or frequency of CPU cores can be reduced to save power. 

The scheduler must update scheduling decisions over time to account for such variability. Moreover, it must be robust to situations where a node is unable to finish processing an input during its allocated CPU time slices.

\begin{figure}[t]
\centering
\includegraphics[width=0.48\textwidth]{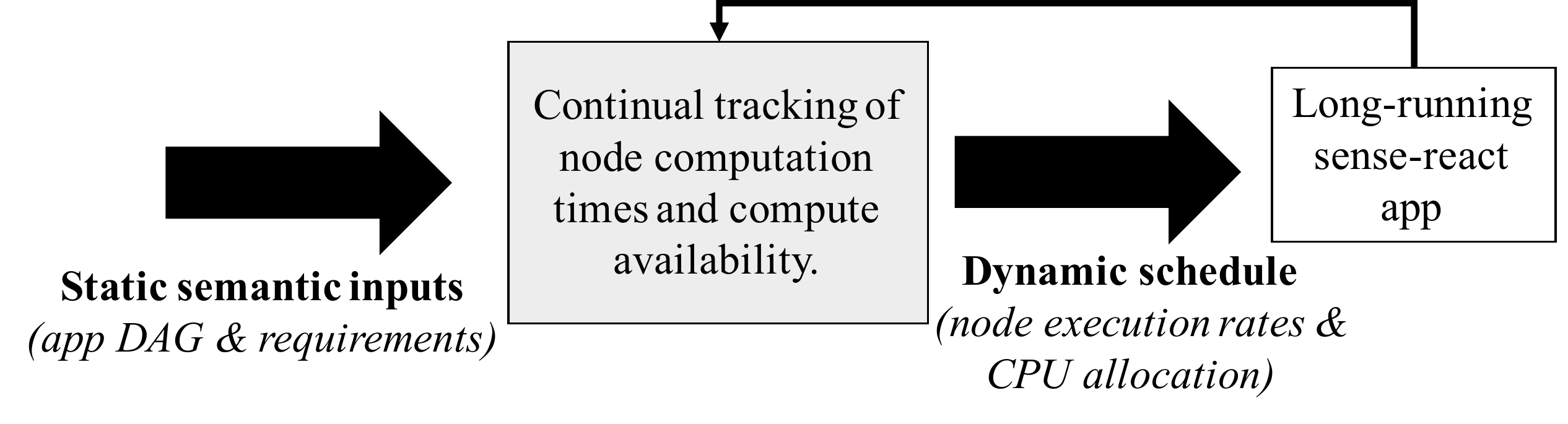}
\vspace{-10pt}
\caption{High-level workflow of \oursys}
\vspace{-15pt}
\label{fig:workflow}
\end{figure}

\subsection{Putting it all together}

%\radhika{better section title}

We design \oursys, a CPU scheduler for \sr application, based on the above considerations. For a given app (represented as a DAG), \oursys makes CPU scheduling decisions (\S\ref{sec:dimensions}) at the granularity of subchains (\S\ref{sec:granularity}), so as to satisfy the semantic preferences and constraints (\S\ref{sec:preferences}) on chain response times and node throughputs (\S\ref{sec:metrics}), while handling potential variations in compute usage and availability (\S\ref{sec:variations}).

%Add figure. Mention what the inputs are more precisely. 

Figure~\ref{fig:workflow} represents the high-level workflow of \oursys. It provides an interface for the app developers to express their semantic requirements. In particular, the developers specify (i) the DAG structure (nodes and edges), (ii) grouping of nodes into subchains, (iii) which metrics must be optimized (i.e. response times of which chains and throughput of which nodes) and the corresponding constraints and weights across these metrics (to capture semantic trade-offs).~\footnote{\oursys allows the developers to optionally specify a few other semantic inputs that we discuss in \S\ref{sec:design}.} \oursys periodically tracks the per-node computation times and compute availability, and adaptively determines the schedule (node execution rates and CPU allocation), so as to best meet the specified semantic requirements. We detail \oursys's design in \S\ref{sec:design}.  

\oursys inputs from app developers require significant domain expertise (in the form of thorough understanding of the app requirements). Current frameworks, on the other hand, require app developers to directly configure the low-level scheduling knobs (node execution rates and system scheduling policies) --- this requires system-level expertise in addition to domain expertise. More importantly, the inputs required by \oursys are largely independent of the compute availability and usage  --- they rely only on app semantics and can remain unchanged across deployment platforms and over time. The precise schedule and system-level knobs, on the other hand, must be dynamically adapted based on varying compute usage and availability (\S\ref{sec:eval} shows how static configurations lead to sub-optimal performance).   
%\radhika{static vs dynamic}
\oursys, thus, simplifies the inputs required from app developers by providing a layer of abstraction that translates \emph{static} semantic requirements into a \emph{dynamic} low-level system configuration and schedule.   

One could simplify the inputs required from app developers even further by adding hooks to the development framework to infer the DAG structure, inferring subchain grouping based on specified performance constraints and weights, and inferring DAG-level performance requirements from app-level performance requirements. These extensions are beyond the scope of this work, and we leave them for future research.

\section{\oursys Design}
\label{sec:design}

%\radhika{update to add some more design details, as per SOSP reviews.}

%\radhika{on some more thought, `Jenga' has a slightly negative connotation that the scheduler is too brittle/fragile. Let's think a bit more. How about we go with `Catan' since we are doing resource management? although we manage only one resource} 

%Start with acknowledging related work -- do not handle all requirements (defer a more detailed discussion of related work to later section). We design a scheduler to handle the requirements. Not the only solution. Can borrow and extend other ideas from the scheduling literature. 

%Can copy most of the old content. Ensure we discuss how each requirement is met, and what the interfaces are. Basically, ensure that the scheduler comes across as a ``common'' scheduling framework that satisfies all requirements. 

%We now describe \oursys, our scheduler designed to handle the above requirements for \sr systems. 

%Note that \oursys is not the only design that can satisfy \sr requirements. System scheduling is a rich field, and while existing schedulers do not directly satisfy all of the above requirements (as we discuss in \S\ref{sec:related}), it could be possible to extend them to meet the requirements. Nonetheless, we believe that \oursys provides a simple and natural starting point that is effective at demonstrating the importance of proper scheduling in \sr systems.

%\subsection{Problem Formulation}
We model our scheduling problem as a constrained optimization problem, wherein the objective is a weighted linear combination of the DAG-level metrics (i.e. chain response times and reciprocal of node throughputs). \oursys schedules the app subchains, so as to optimize the specified objective and meet any specified constraints.\footnote{We allow the constraints to be soft, i.e. if the scheduler cannot meet them, it prints a warning, and aims to meet a looser (scaled up) set of constraints.}

\oursys adopts a hierarchical approach by breaking the scheduling decisions into two stages --- the first stage determines the mapping between subchains and cores (\S\ref{sec:multicore}), and the second stage makes scheduling decisions at per-core and per-subchain granularity (\S\ref{sec:singlecore}). 
We begin with outlining our approach under the assumption that the computation time at each node and the number of cores stays constant, and discuss how we handle variability in \S\ref{sec:dynamicity}.

% \paragraphb{Single Chain Scenario} We would first go over our theoretical result for a DAG with a single chain, which informs our approach for general DAGs. For a single chain C, comprised of nodes $n_1,..n_m$ with compute times $c_1,..c_m$, 
% \aditi{Would it make sense to put the single chain result/formula here?}

% \vspace{-2em}
\subsection{Stage I: Core Allocation}
\label{sec:multicore}
% \aditi{hardness: possible solutions are Exponential in the number of subchain and number of cores}
For a DAG with $N$ subchains on a system with $K$ cores, \oursys's first stage of optimization outputs a boolean matrix of size $N \times K$, where an element $a_{ij}$ is 1 if subchain $i$ is allowed to execute on core $j$. 
%We use $c_x$ to denote the computation time of node x, and $x \in SC_i$ is the set of all nodes which belong to subchain i. 
%We assume that all cores are homogeneous, i.e. equally powerful, but our approach can easily be extended to handle heterogeneous cores \radhika{how?} \aditi{the analytical formulae used in the MILP can be modified to use different compute times for a node based on which core its assigned to. i.e. replace each occurrence of $c_i$ by $\Sigma_{j=1}^k c_i^j * a_{ij} $}.
%\paragraphi{Assumption A1} 
Given the exponential solution space, we add a constraint for tractability: each subchain either runs alone on one or more cores or shares a single core with other subchains (i.e. two or more subchains do not share two or more cores).
We also add a trivial constraint: each subchain should be assigned to at least one core, and vice versa. 

We begin with estimating the period $p_i$ of the subchain $S_i$, as a function of $a_{ij}$ variables.
The period of a subchain captures its execution rate; subchain $S_i$ processes a new input every $p_i$ time units. We consider two cases (based on the above constraints), and make simplifying assumptions in each case for computational tractability -- we relax these assumptions when making finer-grained scheduling decisions in Stage II (\S\ref{sec:singlecore}).  
%the subchain produces a new output every $p_i$ ms.
%  We develop analytical models to compute per-chain latency, throughput, and response times as a function of $a_{ij}$.
%  \aditi{use the following formulae to approximate the period of each subchain as a function of $a_{ij}$}

\paragraphiq{(i)} If multiple subchains share a single core $x$, we assume each subchain gets an equal share of the CPU core (inspired by Linux' fair scheduling ~\cite{linux-cfs, linux-sched}). The period of a subchain $S_i$ that is assigned to core $x$ will then be the sum of computation time of all nodes in $S_i$ multiplied by the number of subchains that are sharing the core $x$.
%, due to our assumption of fair scheduling (which we forgo in \S\ref{sec:singlecore}). 

\paragraphiq{(ii)} If a subchain $S_i$ is assigned $k_i$ cores ($k_i \ge 1$), we assume all nodes in $S_i$ have the same degree of parallelism (i.e. all nodes can be parallelized to use $q$ cores, where $1 \le q \le k_i$) and their compute scales perfectly with it. We use the analytical formulation described later in \S\ref{sec:singlecore} to compute the period $p_i$ based on the computation time of each node in the subchain. 
 %The formulation uses measured computation time of each node. We use our second approximation to tractably model the periods as a function of degree of parallelism ($q_i$) by accordingly scaling the computation time of each node $S_i$ (i.e. if a node $n_ij$ has computation time $c_{ij}$ on 1 core, we assume its computation time on $q_i$ cores is $c^{qi}_{ij} = c_{ij}/q_i$). 
 
%We thus compute periods $p_i$ for each subchain $S_i$ as a function of core allocation $a_{ij}$ and degree of parallelism $q_i$. 

\setlength{\tabcolsep}{14pt}
\begin{table}[]
\begin{tabular}{cc}
% \cline{1-2}
\multicolumn{2}{c}{Chain $C_t$: $S_{t1} \rightarrow S_{t2}..S_{tn}$, with periods $p_{t1}..p_{tn}$}                  \\ \toprule % \cline{1-2}
\multicolumn{1}{c}{\textbf{Metric}}     & \textbf{Approximation}                 \\ \midrule % \cline{1-2}
%\multicolumn{1}{c}{Node throughput}     & Subchain throughput                    \\ % \cline{1-2}
\multicolumn{1}{c}{Chain Latency}       & $p_{t1} + \Sigma_{x=2}^{n} 2*p_{tx}$   \\ %\cline{1-2}
\multicolumn{1}{c}{Chain Throughput}    & $1/(max_{x=1}^{n} p_{tx})$             \\ % \cline{1-2}
\multicolumn{1}{c}{Chain Response Time} & Latency + 1/Throughput 
\\ \bottomrule
\end{tabular}
\caption{Approximate formulae for chain-level metrics. 
Note that we use the worst case latency formula, and average throughput (the exact time gap between two outputs can vary in practice). 
%\radhika{check} \aditi{can vary with time / across different outputs} 
These simple approximations allow us to formulate the problem as MILP and GP in Stages I and II respectively. \vspace{-5pt}}
\label{tab:metric-approximation}
\vspace{-1.5em}
\end{table}

Thus, for a given core allocation $a_{ij}$, we get approximate periods $p_i$ for each subchain $S_i$. We next estimate the chain-level and node-level metrics as a function of $p_i$ for each subchain which, in turn, allows us to estimate the final objective function as a function of $a_{ij}$. For a subchain $S_i$ with period $p_i$, the throughput of each node in $S_i$ is equal to the subchain's throughput, i.e. $1/p_i$. Table \ref{tab:metric-approximation} lists how we estimate chain-level metrics, for a chain $C_t$ comprising of subchains $S_{t1} \rightarrow S_{t2} \rightarrow S_{t3}..S_{tn}$. The average chain throughput is bottlenecked by its slowest subchain. We estimate the worst-case chain latency as follows: at each subchain $S_{ti}$ in the chain, a new output $o_{t(i-1)}$ from the preceding subchain ($S_{t(i-1)}$) might have to wait for a whole period, (i.e. $p_{ti}$) for $S_{ti}$ to finish its current execution (except for at $S_{t1}$ where there's no waiting time), and it will take $p_{ti}$ time for $S_{ti}$ to fully process $o_{t(i-1)}$ and produce a new output.~\footnote{It is a common practice in real-time literature ~\cite{davare2007period} to compute worst case latency of chains in this manner.} We then estimate the chain response time as the sum of the chain latency and period (reciprocal of throughput).

 Using the models above,
 we formulate a Mixed Integer Linear Program (MILP) that solves for $a_{ij}$
%  \radhika{removed $q_i$ -- to discuss}
 %and $q_i$ \footnote{Note that only the optimal values of $a_{ij}$ are used, and we solve for the best degree of parallelism for each subchain in \S\ref{sec:singlecore} by forgoing the assumptions of this stage.} 
 such that the specified objective function is optimized. Intuitively, the solver determines the right trade off of which subchains can share a single core vs which ones should be allowed to scale up to multiple cores, based on the specified weights and constraints.

\subsection{Stage II: Per Core and Per Subchain Scheduling}
\label{sec:singlecore}
Given the subchain to core mappings from our first optimization stage, the next stage makes finer-grained scheduling decisions for each subchain and core. We consider two cases: 

% \begin{algorithmic}
% \State $p_{opt} \gets +inf$
% \State $q_{opt} \gets 0 $
% \For{q in [1,k]}
%     \State $p_q = max( max (), \Sigma () / k/q )$
%     \If{$p_q\leq p_{opt}$}
%         \State $p_{opt} = p_q$
%         \State $q_{opt} = q$
%     \EndIf
% \EndFor
% %     \If{$i\leq 3$}
% %         \State $i \gets i+2$
% %     \EndIf
% \end{algorithmic}

% \begin{algorithm}[ht!]
% \SetAlgoLined
% % \KwResult{$q_i$}
%  $q^*_i = 1, q = 1, rt^*_i = \infty, p^*_i = \infty$\;
%  \For{q in [1,k]}{
%  $p_q = max(max_{j=1}^m (c^q_{ij}), \Sigma_{j=1}^{m} (c^q_{ij})/(\lfloor k/q \rfloor))$
%   $rt_q = \Sigma_{j=1}^m (c_{ij}^q) + p_q$\;\\
% %   freq = 1.0/(max(max_{j=1}^m (c^q_j), \Sigma_{j=1}^{m} (c^q_j)/numBins))\;\\
%   \If{$rt_q \leq rt^*_i$}{
% %   bestF = freq\;
%   $rt^*_i = rt_q$\;
%   $q^*_i = q$\;
%   }
%  }
%  \caption{Algorithm for choosing the best degree of parallelism ($q^*_i$) and period ($p^*_i$) for a single subchain $S_i$ on k cores}
%  \label{alg:single-chain-theory-opt-q}
% \end{algorithm}

\paragraphb{Single subchain on one or more cores} For a subchain $S_i$ comprising of nodes $\{n_{i1}, n_{i2}, \ldots n_{im}\}$ with $k_i$ assigned cores, the scheduler computes the time period $p_i$ of the source node ($n_{i1}$) and the degree of parallelism for a node in the subchain ($q_i$), such that the response time along the subchain is minimized. 
%(which, in this case and under our theoretical assumption of constant computation times, would simultaneously maximize throughput and minimize latency). 
% Q. DO we choose the best RT or the best period? A. they have the same optimal, atleast for single chain single threaded case. Not fully sure if true for 1chain multi-threaded case.

The period of the subchain corresponding to a given value of $q \in [1,k_i]$ is given by:% \aditi{We iterate over all possible values of q from 1 to k, where in, for each }
\vspace{-5pt}
  \begin{equation} \label{eq:single-chain-theory-period}
     p_i(q) = max(max_{j=1}^m (c^{q}_{ij}), \Sigma_{j=1}^{m} (c^{q}_{ij})/\lfloor k_i/q \rfloor)
 \end{equation} 

\noindent where $c^{q}_{ij}$ is $n_{ij}$'s computation time with at most $q$ cores (only a subset may be designed to use all $q$). Note that, unlike period computation in Stage 1, we allow different nodes in the subchain to use different degrees of parallelism in this stage.
% ~\footnote{Our period computation for multi-core allocation in Stage 1 uses Eqn\eqref{eq:single-chain-theory-period}, under the assumption that all nodes use exactly $q$ cores. We relax that assumption in this stage.}
% , and set $p = max(max_{j=1}^m (c^q_j), \Sigma_{j=1}^{m} (c^q_j)/\lfloor k/q \rfloor)$, where $c^q_j$ is $n_j$'s computation time with at most $q$ cores. 
%Intuitively, the period of a subchain is bottlenecked by its slowest node if $max_{j=1}^m (c^{qi}_{ij}) > \Sigma_{j=1}^{m} (c^{qi}_{ij})/\lfloor k/q_i \rfloor$, and by the available compute capacity otherwise. 
%As described in Algorithm \ref{alg:single-chain-theory-opt-q}, 

The corresponding response time of the subchain, for a given $q$, can be analytically computed as $(\Sigma_{j=1}^m (c_{ij}^{q}) + p_i(q))$.
We iterate over all possible values of $q \in [1,k_i]$ and select the value $q_i$ (and the corresponding $p_i$) that results in lowest response time for the subchain.
%For each value of $q_i$, we allow each node in the subchain to use at most $q_i$ cores (only a subset may be designed to use all $q_i$), and $c^{qi}_{ij}$ denotes the node $n_{ij}$'s compute time with atmost $q_i$ cores. \radhika{removed algo -- discuss}
%Note that a larger $q_i$ may reduce per-node processing times (since $c^q_{ij}$ decreases with increasing $q_i$), but it also reduces the degree of pipelining along the subchain (i.e. the numbers of nodes that run at the same time, since each node is using more cores) \radhika{cut?}.
%The scheduler 
We have proved that this rate allocation achieves subchain response time within $2 \times$ the optimal, and equal to the optimal if all the nodes in the subchain can only use a single thread (i.e. $q_i = 1$) ~\cite{supplementary-proofs}.
% for $q = 1$. 
% The value of $q$ ranges from $1$ to $k$; 

% We select the value of $q_i$ that results in the lowest response time, as approximated in the above algorithm.

%We shall now extend the approach mentioned in \S\ref{sec:multicore} (Case 1). Considering a subchain of nodes $\{n_1, n_2, \ldots n_m\}$ with $k$ assigned cores. The scheduler must determine the rate (or the time period $p$) of the source node ($n_1$) and the degree of parallelism for each node, such that the response time is minimized. We allow each node in the subchain to use at most $q$ cores (only a subset may be designed to use all $q$), and set $p = max(max_{j=1}^m (c^q_j), \Sigma_{j=1}^{m} (c^q_j)/\lfloor k/q \rfloor)$, where $c^q_j$ is $n_j$'s computation time with at most $q$ cores \footnote{Note that we are not assuming perfect scaling here, and can use the actual/empirical computation time of each node given q cores}. We have proved that this rate allocation achieves response time (for this particular subchain) within $2 \times$ the optimal, and equal to the optimal for $q = 1$~\cite{supplementary-proofs}. The value of $q$ ranges from $1$ to $k$; a larger $q$ may reduce per-node processing times, but also reduces the degree of pipelining along the subchain. The scheduler iterates over all $k$ choices for this subchain, and selects the value of $q$ that results in lowest response time for this subchain.

\paragraphb{Multiple subchains on a single core} For subchains  $S_1, S_2 ...S_n$ assigned the same core, the scheduler must determine their periods and temporal allocation of CPU time across them. 
%We define a variable $f_i$ for each subchain, which denotes two things: a) the subchain gets $\Sigma_{x \in SC_i} (c_x) * f_i$ CPU time per period, b) the subchain finishes one execution per $1/f_i$ periods. The total period of this schedule then comes out to be $\sum_{i \in C_j} f_i * (\Sigma_{x \in SC_i} (c_x))$. 
We construct a periodic schedule, and execute $f_i$ fraction of subchain $S_i$ in each period, such that $\Sigma_{i=1}^n (f_i * c(S_i))$ equals the period of the schedule, where $c(S_i)$ is the sum of computation time for all nodes in $S_i$. 
%(we assign $f_i$ such that $1/f_i$ is integral). 
The execution of each (fractional) subchain within a period follows a configurable ordering.~\footnote{By default, \oursys assigns the order based on specified weights, but provides an option to directly specify it as a semantic input.}
%By design, the output of such a schedule ensures that the CPU core is fully-utilized. If, however, the  
%~\footnote{We can generalize our solution to not use the full CPU if the robot needs to save battery life and the required performance can be achieved with less than a core : add fake node}, while 
%The fraction variables tune the core allocation across subchains. 
Each subchain will finish processing one input once every $1/f_i$ periods, i.e. $p_i = \frac{\Sigma_{i=1}^n (f_i c(s_i))}{f_i}$. We allow $f_i$ to be larger than 1 for subchains that 
%\radhika{/*} exhibit high variability in computation times across inputs and \radhika{*/ cut?} 
require optimizing \emph{average} throughput (e.g. \gml in robot navigation)~\footnote{We cannot assume a fixed time period of input processing for nodes that buffer and batch process inputs, and instead consider their average throughput. Such nodes can either be auto-identified (given support from framework) or can be explicitly identified by the developer.}, and constrain $1/f_i$ to be an integer for other subchains to allow the scheduler to control the exact throughputs.
%We distinguish between subchains (or nodes) which have constraints on average throughput (e.g. streaming nature such as GML) from others, by allowing f to take arbitrary values for such nodes, but constraining it to be integral - reciprocal (i.e. $1/f_i$ is an integer) for other nodes. This is because, $1/f_i$ being integral allows the scheduler to control the exact throughput of the subchain.
We combine the above period formula with metric estimations given in Table \ref{tab:metric-approximation}
% We approximate the low-level metrics as an analytical function of these fraction variables \aditi{add formula $p_i = \frac{\Sigma_{j=1}^n (f_j * c(s_j))}{f_i}$}
, and formulate a Geometric Programming problem to compute the $f_i$ and $p_i$ variables such that the specified objective function is optimized under the specified constraints. 
\subsection{Handling Variations}
% \vspace{-0.25em}
\label{sec:dynamicity}

\paragraphb{Coarse Grained Variations} We handle coarse-grained variations over time by continually recording the computation time across all nodes and tracking the number of available cores. We re-compute the stage II scheduling decisions (\S\ref{sec:singlecore}) periodically using the 95\%ile computation time for each node measured over the previous 50 values in our implementation. 
% ~\footnote{We use average for nodes that require average throughput control e.g. \gml} 
We handle multimodal computation times by taking the weighted sum of 95\%ile computation times across the different modes. 
%\radhika{how do we determine different modes? - input from application for now, it seems non-trivial to infer modes from a distribution.}. 
We invoke the more expensive stage I optimization (\S\ref{sec:multicore}) less frequently.
%-- once every 20s in our implementation. 
%We chose these periods so as to keep the overhead of re-solving low.

\paragraphb{Fine Grained Variability} 
%\radhika{Aditi, please add in the text for what you are doing here}
%Since the solvers cannot handle sudden variations in compute requirements of nodes, 
In spite of periodically adapting the scheduling decisions, a node may still exceed its expected (previously measured) computation time. To handle these situations, the scheduler implements a priority-based stealing mechanism, wherein if nodes $A$ and $B$ share a core, and $B$ has lower priority than $A$, then in each period of the schedule, the scheduler allocates $B$'s CPU time to $A$ if the last output from $A$ was not received at its expected time period.
%\radhika{this sentence no longer makes sense to me}. 
We infer node priorities from the specified weights, and also provide the option to explicitly specify these as semantic inputs.
\section{Implementation}
\label{sec:impl}
We implement \oursys scheduler on top of Linux, as a ROS node for robotics and as an ILLIXR plugin for VR, both of which use the same backend which handles all the scheduling decisions. 

\paragraphb{Initialization} 
%\oursys assumes that each node runs in a thread of its own. 
%\radhika{add 1-2 lines on how the app DAG is provided as an input to \oursys}
% \aditi{dag input: frameworks have provision for the name of each node, each node sends their thread ids} \oursys is initialized with an input file containing the DAG structure. 
At initialization time, \oursys requires the DAG structure and the thread ids corresponding to each node, 
since they are required to enforce fine grained scheduling. Note that both of the above inputs can be automatically obtained by adding hooks to the framework. For robotics, we modify the ROS communication library~\cite{ros-comm} to expose the ids of all the threads it uses under-the-hood to handle communication between nodes. 

\paragraphb{Bootstrapping} \oursys takes the application DAG and constraints/weights as input, and bootstraps the core allocation by assigning equal compute time to all the nodes and running the Stage - I solver.~\footnote{
We set same compute times of all nodes at bootstrap, so as to share the CPU equally among all nodes (the exact value, set to 5ms, does not matter).} Based on the output, \oursys spawns a thread $TQ_j$ for each core with multiple subchains, and a thread $TS_i$ for every subchain assigned to one or more cores, to handle the per-core and per-subchain scheduling respectively. It bootstraps the per core fractional schedule by assigning a fraction of 1 to all subchains. The bootstrapped configuration lasts only for the first 2s, until actual node computation times are measured and the solver is invoked. Lastly, it spawns a dynamic re-optimization thread which periodically updates the scheduling decisions. All the scheduler threads ($TQ_j$ and $TS_i$) are assigned the SCHED\_FIFO policy with a very high priority of 4, except for the re-optimization thread which runs with the SCHED\_OTHER policy, so as not to interfere with the application's execution.

\paragraphb{Runtime} At runtime, each scheduler thread $TS_i$ triggers the first (source) node of the subchain $i$ at the analytically computed time period. The threads $TQ_j$ enforce the core $j$'s periodic fractional schedule using two mechanisms. First, 
it triggers the first node in each subchain $s_y$ once every period if $f_y \ge 1$, and
%the subchain buffers inputs (or has high variability in compute times e.g. GML) and 
once every $1/f_y$ periods otherwise. Secondly, in each period of the schedule, it assigns SCHED\_FIFO policy with the lowest priority of 1 to all the node threads, and iteratively bumps up the SCHED\_FIFO priority (to 2) of (all the threads of) each subchain for the computed fraction of its time within the period. As described in \S\ref{sec:dynamicity}, the controller can enforce priority-based stealing within each period. For threads that we do not schedule explicitly,
(e.g. IMU in ILLIXR which needs to run at high frequency and has negligible compute), 
we assign SCHED\_FIFO policy with a fixed high priority (3) so that they can preempt any node whenever needed~\footnote{One could use Linux' SCHED\_DEADLINE scheduler to implement \oursys's scheduling decisions instead of SCHED\_FIFO, but we found it to be too inflexible for handling variations.}.

The scheduler takes 0.12ms to pause the execution of one node, and start executing the next. We add an extra constraint on each subchain $c(s_i) * f_i \geq 1$ to limit the scheduler's intervention (and the ensuing overhead due to that). 
%\radhika{cut /*} 
% This limits the period of the CPU core to be atleast the number of subchains executing on the core (in ms). We can extend the fractional policy by allowing only a subset of nodes/subchains to run each hyper period, so as to support faster periods with less overhead. 
%\radhika{*/} 
We also add 5\% slack time in each period, which can be used by the re-optimization thread or the application threads.\footnote{This is done to account for the fact that Linux does not allow SCHED\_FIFO processes to use more than 95\% CPU time ~\cite{linux-sched}.} We scale the analytical functions for the low level metrics accordingly, in both the optimization formulations.
% imu - int prio=3 always.

\paragraphb{Dynamic Re-optimization} We periodically re-solve for the new optimal scheduling decisions, based on the latest compute time estimates of all nodes (as discussed in \ref{sec:dynamicity}). We have implemented both the MILP and GP formulations using the Mosek Fusion C++ library. It takes our solver 24-26ms to solve the GP for the navigation application and 5-6ms for VR. Solving the MILP is more expensive, requiring 60ms for the navigation DAG for 2 cores. We invoke both the solvers 2s after initialization, and then re-run Stage I every 20s, and Stage II every 5s. Increasing the time periods can decrease the overhead of running the solvers. We chose the periods so as to keep the overhead roughly under 0.05 CPU cores\footnote{The process of selecting the solver periods can easily be automated based
on this criteria by tracking the solvers’ compute usage.}. For each execution of the Stage I optimization, if the core allocation changes, then we kill the existing $TQ_j$ and $TS_i$ threads and re-initialize them based on the new solution. We kill the threads for simplicity, and observe negligible overheads due to this, but the implementation can be updated to send information to the existing threads instead of killing them.

% \radhika{How different are ROS and Illixr scheduler implementation? Does this common implementation section make sense?} \aditi{same solver, implementation differs in taking render order as input, and waiting for render to finish}

% \radhika{this should also talk about how you are assigning threads to cores, and how you are explicitly controlling the node executions in fractional schedule.}

% \radhika{also mention how scheduler supports nodes that are not explicitly scheduled --e.g. IMU in VR and odometry in navigation.}

\section{Evaluation}
\label{sec:eval}

%We now evaluate how \oursys influences the performance of the three applications described in \S\ref{sec:background}. 

\subsection{Face Tracking}
\label{sec:facetracking-cs}
% \aditi{Explain default = hand tuned frequencies chosen by the respective application developers, possibly based on intuition, with other sched decisions left to the OS}
% \aditi{Used 1-2cores: common for robots to have lightweight compute ~\cite{p3at} has a dual core CPU }
\begin{figure}[t]
\includegraphics[width=0.49\textwidth]{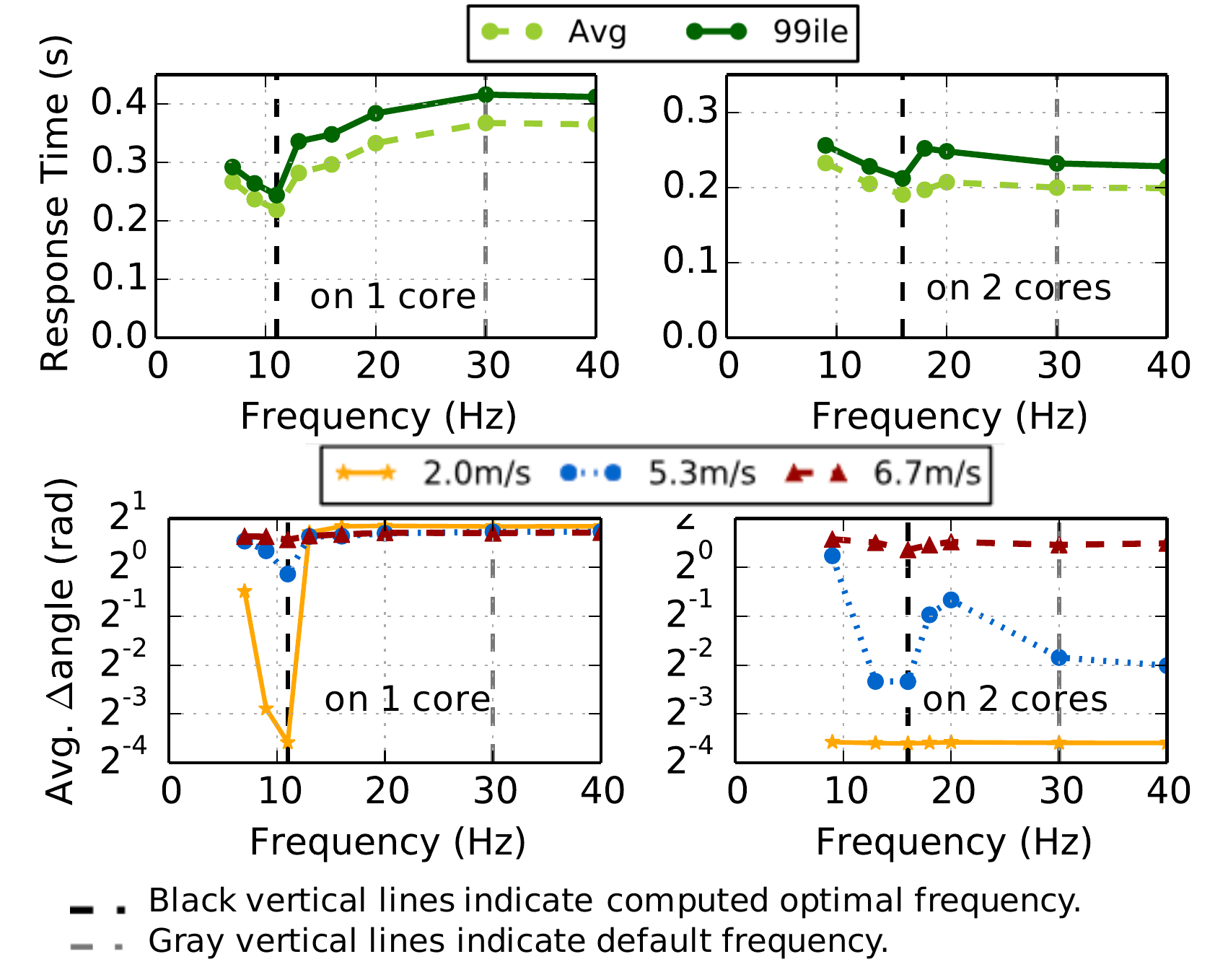}
\vspace{-20pt}
\caption{Response time (top) and tracking performance for different object speeds (bottom) as source frequency is varied.}
%Notice different y-axis range for \rt plots, and log-scale y-axis for $\Delta$angle.}
\label{fig:facetrack}
\vspace{-10pt}
%\hrule
%\end{figure}
\end{figure}

%Application overview

Our first study highlights the importance of scheduling decisions in \sr systems in the absence of complex semantic trade-offs and variability.  

\paragraphb{Scheduler Configuration}
%Since there is only a single chain (that we model as a single subchain), the scheduling problem boils down to the case of a single subchain on one or multiple cores (from \S\ref{sec:singlecore}). 
As the app comprises of only a single chain, the scheduling objective is to minimize its response time. All app nodes belong to the same subchain.  

\paragraphb{Experiment Setup}
We use Gazebo~\cite{gazebo} (a standard physics simulator used for evaluating robotics applications) to simulate a dummy moving along a square path centered at a robotic camera that can rotate at a velocity of up to 3.2rad/s (\cite{tb3-actuator-speed-1, tb3-actuator-speed-2}). Gazebo feeds camera inputs to detection and planning nodes implemented in ROS, which in turn feed the velocity output back to the simulated camera. 
%We make a few modifications to this basic setup to better model realistic timings. 
 
 We run the whole setup (i.e. the simulator as well as the application) on an AWS EC2 instance (m5.4xlarge). Gazebo runs on 5 reserved cores, and sends camera inputs as fast as it can to a shim source node, which models the time taken for post-processing or formatting camera inputs (measured as 25ms using the usb-cam module \cite{ros-usb-cam}) and then publishes the latest input at the specified frequency. 
%  We add a node \aditi{this is the shim node!?} that .
 Since the algorithm to detect the simulated dummy is trivial, we model realistic timings by augmenting our detection node -- every time it processes a simulated frame, it also runs a single-cascade HAAR algorithm~\cite{haar, ros-face-tracking} on a frame drawn from a real video data-set \cite{vw-realvideo-dataset, vw-realvideo-dataset2}  (the time taken for this varies between 55-60ms). The planning node (which takes 1ms) tracks the dummy using the detection output from the simulated frame. We experiment with varying source frequency and speed at which the dummy moves, as well as with different number of cores $k$ made available to the application (that excludes Gazebo).

\paragraphb{Baselines} We compare \oursys to the default configuration, which is based on the source frequency hand-tuned by the application developer (30 Hz) and uses Linux' default policy SCHED\_OTHER (based on Completely Fair Scheduler) for all other scheduling decisions. This configuration remains the same regardless of the number of cores.

\paragraphb{Metrics} We evaluate \oursys on the chain response time as well as the robot's face tracking performance. We measure the tracking performance as average angular distance ($\Delta$angle) between the camera's orientation and the position of the dummy (a lower value implies better performance).

\paragraphb{Results} 
 Figure~\ref{fig:facetrack} shows the response time and tracking performance for a range of source frequencies on a system with one and two cores (to model platforms with limited compute). The black and gray vertical lines indicate the optimal value assigned by \oursys, and the default configuration respectively.
 %The optimal here is based on our theoretical analysis for a single chain DAG which minimizes the chains' response time. 
 The compute capacity is the bottleneck with one core, while the slowest node is the bottleneck with two cores.
 %The top graphs in Figures~\ref{fig:facetrack} reveal that 
 
 Response time (shown in the top graphs) is lowest at the optimal source frequency computed by \oursys using the analytical formulation in \S\ref{sec:singlecore}. This is because a lower than optimal frequency lowers throughput of the chain, while higher than optimal frequency increases latency (detailed explanation omitted for brevity).  
 %At frequencies higher than optimal, the amount of wasted work increases -- the first two nodes now process extra inputs that get dropped at the slowest node. This has two effects. First, since the processing rate at the slowest node can no longer keep up with input arrival rate, the latest input must wait in the queue (of size 1), increasing the latency (and thus response time). With two cores, the increase in latency due to this effect is highest just beyond the optimal frequency -- even higher frequencies reduce the waiting time by allowing newer inputs to replace the older one (in the queue) at faster rates. The second effect is seen with one core, where the increased work at higher frequencies causes greater contention between nodes for limited compute resources, which further increases latency.
 
 %A higher than optimal value increases its latency , Both of these effects  influence tracking performance.
 We draw the following key observations with respect to the tracking performance (shown in the bottom graphs): (i) Tracking performance strongly correlates with the response time, confirming the impact of low-level metrics on app performance. (ii) The system has the best tracking performance when running at the optimal frequency configured by \oursys, outperforming the default configuration. This shows the impact of scheduling on app performance. (iii) In general, it is harder to track a faster moving object. \oursys's optimal frequency is able to track the object for wider speed ranges.

\subsection{Robot Navigation}
\label{sec:navigation-cs}
\paragraphb{Scheduler Configuration}
We model the objective function as the weighted sum of response times for the different chains involving each node in the navigation DAG (detailed below), and the average output period for \gml (i.e. the average time interval at which it processes scans) as a measure of its throughput. 
We do not explicitly schedule odometry, and configure Scan $\rightarrow$ \lcm $\rightarrow$ LP as a single subchain. All other nodes shown in Figure~\ref{fig:nav2d-dag} (i.e. \gml, \gm, \gp, and \nc) are treated as individual subchains, that can run at different rates. 
%\radhika{scan is not quite part of the same ``subchain'' right? since it runs at a frequency lower than LC and LP?}. \radhika{list other subchains (GML, MP, NC, NP). Also say we do not explicitly schedule odometry (and scans?)} 
%We combine the response times along the chains Odom $\rightarrow$ LC $\rightarrow$ LP, Odom $\rightarrow$ LP and Scan $\rightarrow$ LC $\rightarrow$ LP into a single term (since their analytical value is the same due to both scans and odometry being published at 50Hz by the simulator \radhika{to discuss -- why will 50Hz matter here?}) 
The analytical response times for the local chains (involving the \lcm and \lp) sourced at scans and at odometry differ only by a constant (odometry's period) -- we combine them into a single response time metric for our objective function, assigning it the highest weight of 1.0 (as they are responsible for avoiding collisions). We use a weight of 0.5 for \gml's average output period (to ensure freshness of pose estimates and odometry for path planning). In addition to these, our objective function includes the response time along three chains that  include the three remaining processing nodes (\gm, \gp, \nc) sourced at \gml -- each assigned a small weight of 0.005 (to ensure that these nodes are not starved as the local chain and \gml are prioritized).~\footnote{When analytically computing response times along the chains sourced at \gml, the scheduler uses the empirically observed value of the rate at which it outputs filtered scans (1.5Hz at 75\%ile).} 
%The scheduler can be extended to dynamically measure this rate and change the corresponding constraint at runtime.}
%Since GM only executes if there is atleast one new filtered scan, we enforce the constraint on GM's rate to be slower than 1.5Hz.
%When analytically calculating the response time along any chain that uses GML's filtered scan outputs, we use 1.5Hz as GML's period based on this empirical measurement (the scheduler can be extended to dynamically measure this rate and change the corresponding constraint at runtime).}
We also add the following application specific constraints into our optimization formulation: (i) \gml's average throughput must be at most 50Hz, based on the maximum frequency at which the simulator can publish laser scans. (ii) \gp's throughput must be at most 1Hz. 

%The scheduler is configured to support priority-based stealing for nodes in the local chain and for \gml. 
%The local chain nodes (LC and LP) have the highest priority are never preempted if they overrun. Additionally, \gml has higher priority than GM and GP, allowing \gml to steal CPU time from these two nodes if it overruns. 

\begin{figure}[t]
\centering
\setlength{\tabcolsep}{1pt}
  \renewcommand{\arraystretch}{0.9}
  \resizebox{\linewidth}{!}{
   \begin{tabular}{cc}
    % \includegraphics[width=0.2\textwidth]{cs_thesis_latex_template-2020/figures/GM_run1.pdf}
    % &
     \includegraphics[height=0.14\textwidth]{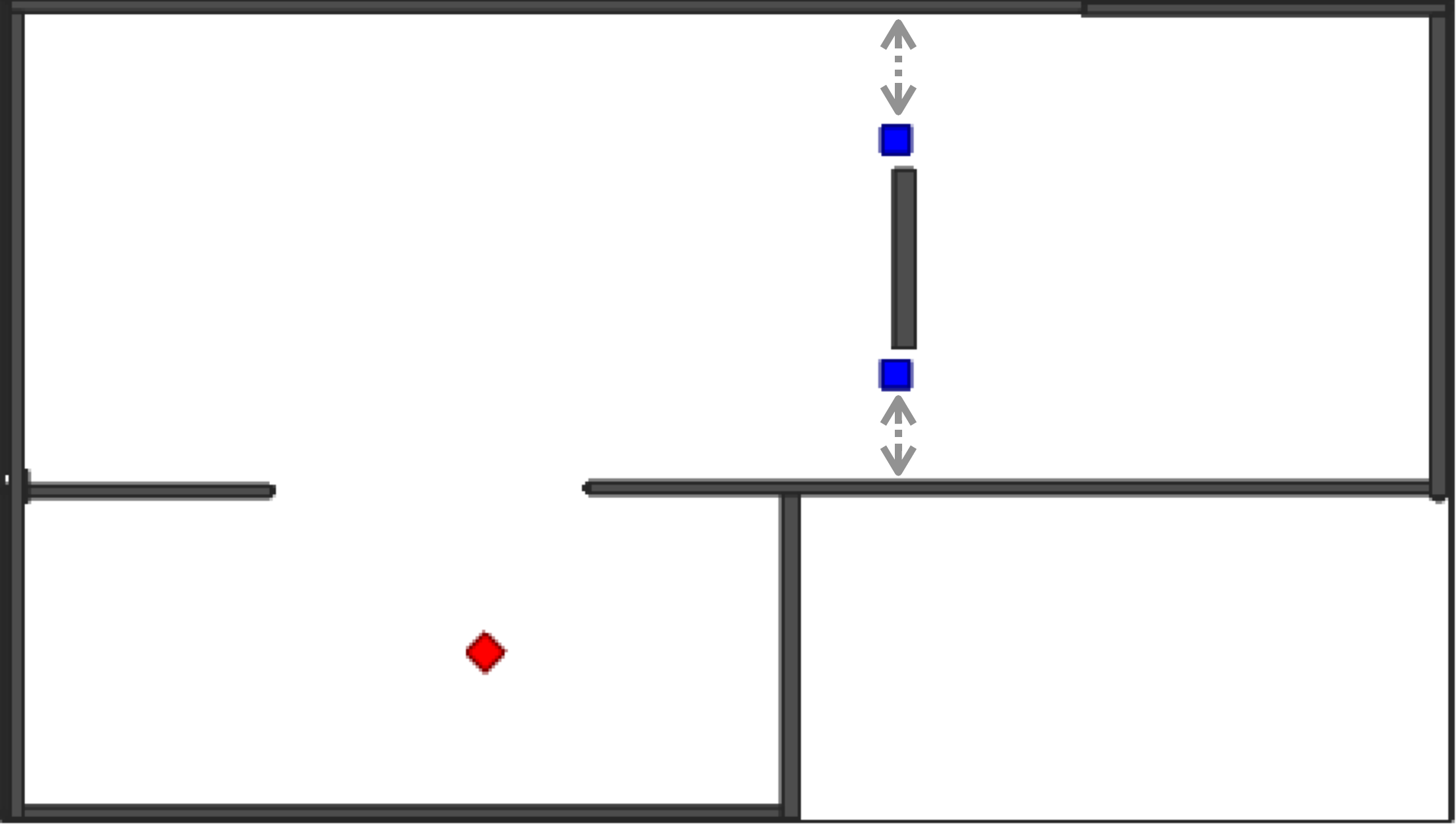}
     &
     \includegraphics[height=0.14\textwidth]{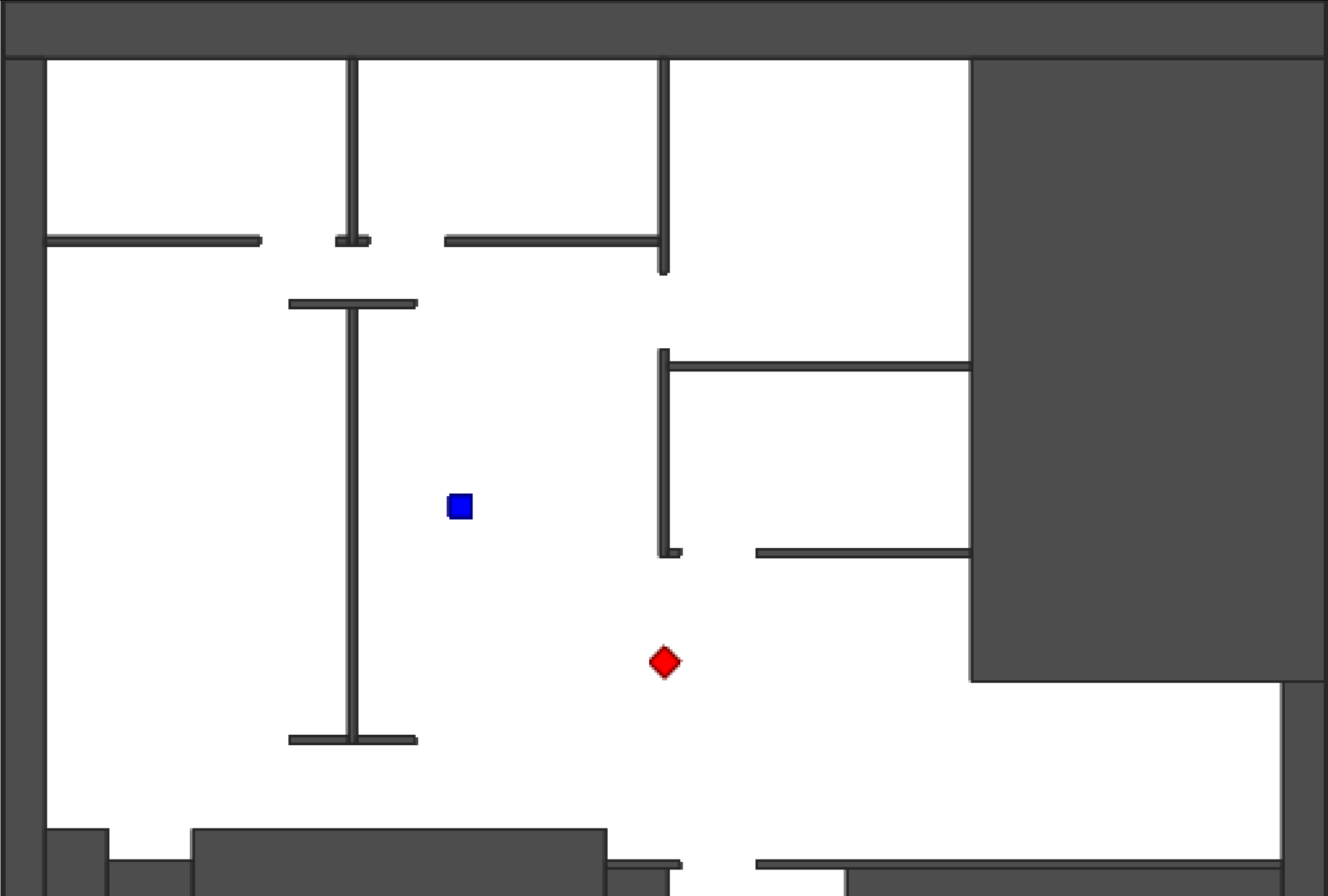}
    
    \end{tabular}}
    \caption{Map 1 (left) and Map 2 (right) used for navigation experiments. The blue squares represent the obstacles and the red square represents the robot. The obstacles in Map1 are dynamic, and move along the two doorways.}
    \vspace{-10pt}
    
\label{fig:nav2d-maps}
% \vspace{-10pt}
\end{figure}

\paragraphb{Experiment Setup}
We use Stage~\cite{Stage} to simulate a P3AT robot \cite{p3at}. The simulator feeds laser scans and odometry into the navigation application, which is implemented on top of ROS~\cite{nav2d}. The application feeds the robot's velocity back into Stage. We configure the simulator to publish laser scans and odometry at 50Hz. %(we have sped up simulation by 5x).
% \aditi{The laser scans are sent to a shim source node, which publishes the latest scan at }
We conduct experiments on two maps: (i) Map 1 (Figure~\ref{fig:nav2d-maps}(left)) requires the robot to move around a dynamic obstacle at the entrances of a room (this models realistic scenarios where collisions with dynamic obstacles may occur at narrow doorways). This map allows us to evaluate the ability of the robot to avoid collisions with dynamic obstacles under different configurations. (ii) Map 2 (Figure~\ref{fig:nav2d-maps}(right)) requires the robot to explore a larger area. 
%We use this map to evaluate other metrics (i.e. the area exploration rate and the staleness of robot's position used for generating navigation commands). 
We do not add any dynamic obstacles to this map, focusing on other performance metrics. Unless otherwise specified, all nodes run on a system with one core (it is common for robots to have single / dual core on-device compute~\cite{p3at}).
%For both of these maps, the robot starts at a fixed location, and needs to explore the whole area, while avoiding collisions with the obstacles. 
%We do multiple runs for each scheduling configuration (\oursys and baselines). %where a run starts with the robot at the fixed location, and ends if the robot is either (a) unable to plan a path from its current location (e.g. stuck near an obstacle), or (b) has no unexplored areas based on the GM's map (which might not be accurate).
% \aditi{ Note that a collision can lead to any of a,b.}

%Obstacle map - mention that it represents a general scenario with dynamic obstacles, e.g. opening/closing doors. [cites?]

%\paragraphb{Metrics} We consider the following key application metrics -- the number of collisions, the rate of exploration, and the staleness of odometry used by NC.

\begin{figure}[t]
\includegraphics[width=0.47\textwidth]{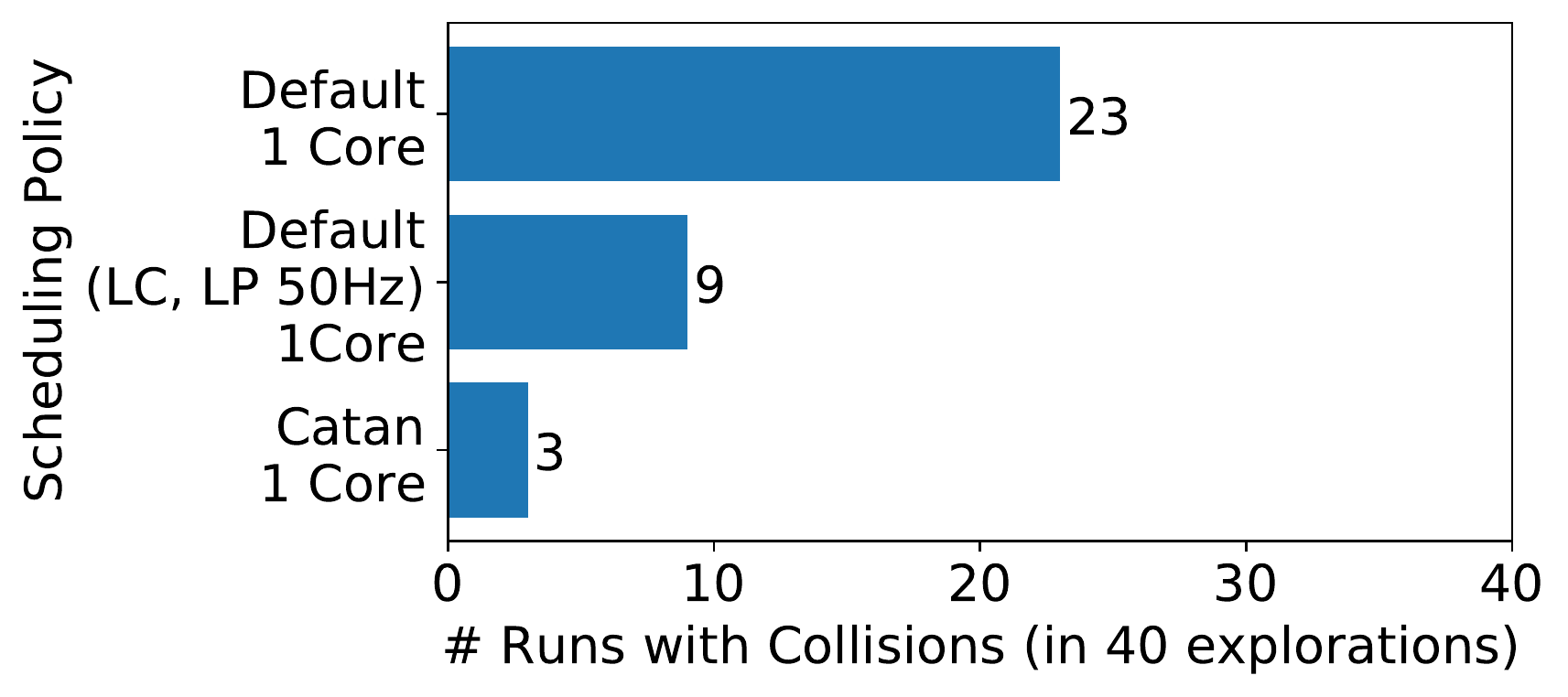}
\vspace{-10pt}
\caption{No. of runs with collisions (out of 40) on Map1, with (i) Default config, (ii) Default with LC, LP at 50Hz, and (iii) with \oursys, 
%all three running on a single core.
\vspace{-5pt}}
\label{fig:nav2d-smallmap-collns-def-dyn}
\vspace{-7pt}
%\hrule
%\end{figure}
\end{figure}

\begin{figure}[t]
\centering
%\setlength{\tabcolsep}{1pt}
%  \renewcommand{\arraystretch}{0.9}
%  \resizebox{\linewidth}{!}{
%   \begin{tabular}{cc}
    % \includegraphics[width=0.2\textwidth]{cs_thesis_latex_template-2020/figures/GM_run1.pdf}
    % &
     %\includegraphics[width=0.24\textwidth]{figures/NO_XGTArea_Avg_1c_Def_Dyn.png}
     %&
     \includegraphics[width=0.37\textwidth]{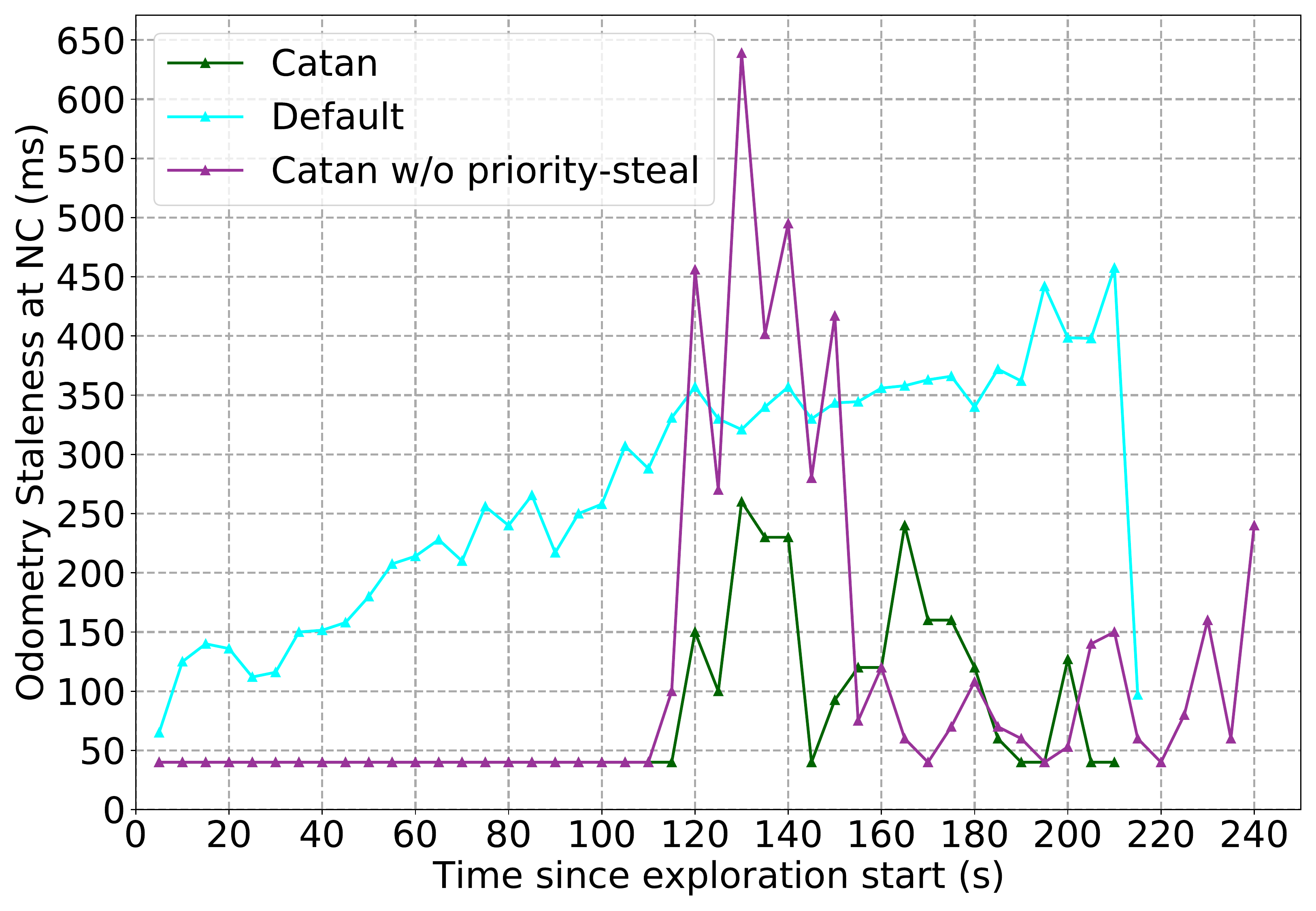}
    
%    \end{tabular}}
    \caption{Tail odometry staleness at NC 
    on Map2 for \oursys vs Default vs \oursys w/o priority stealing (median over 20 runs).\vspace{-5pt}} 
\label{fig:nav-lat-default-wf}
% \vspace{-2em}
\end{figure}

\paragraphb{Comparison with manually-tuned configuration}

\paragraphi{Baselines} We begin with comparing the performance of \oursys with the default configuration (manually-tuned by the developer of the navigation app). In line with the interface currently exposed by ROS, the default configuration only specifies the execution rate of each node~\cite{nav2d} (set to 5Hz for \lcm, 10Hz for \lp, 0.2Hz for \gm, upto 1Hz for \gp, 10Hz for \nc), relying on Linux' default policies (SCHED\_OTHER based on Completely Fair Scheduler) for other scheduling decisions.

\paragraphi{Metrics} We evaluate \oursys against the default configuration on the robot's performance for both Map1 and Map2. For Map1, we do 40 experiment runs and compare the number of runs in which the robot collides with an obstacle. For Map2, we compare the staleness of odometry used for generating navigation command and the area exploration rate. We capture odometry staleness by recording the time difference between an odometry reading and when the corresponding pose is used by \nc. For each experiment run, we aggregate the staleness values collected over time buckets of 5s by computing the 95\%ile values. For each 5s bucket, we report the median of these values across 20 runs. We define the area exploration rate as the time taken to explore 90\% of the map area, averaged over 20 runs. We omit the comparison of low-level metrics for brevity, but \oursys is able to achieve better response time for all chains.

\textit{Map1 Results.} Figure~\ref{fig:nav2d-smallmap-collns-def-dyn} compares the number of experiment runs (out of a total of 40) in which the robot collides with an obstacle on Map1. With the default configuration (which runs the \lcm and \lp at low rates of 5 Hz and 10 Hz), more than 50\% runs suffer from collisions. Increasing the rate of \lcm and \lp nodes to 50 Hz reduces the number of collision to 9. With a high weight assigned to the response time along the local chain, \oursys is able to run the local chains at an even higher rate (125 - 170 Hz), reducing the number of collisions to only 3. These results highlight the impact of scheduling decisions on a robot's performance. 

%A natural question that arises is whether \oursys' ability to better avoid collisions comes at the cost of a large compromise on other metrics. 

\textit{Map2 Results.}
Figure~\ref{fig:nav-lat-default-wf} shows the odometry staleness metric. We find that, in general, \oursys has lower odometry staleness than the default policy. When \gml's compute usage spikes up (especially during loop closures) and as the compute load increases over time (due to increased CPU usage for GM and GP), \oursys is able to explicitly prioritize \gml over GM and GP due to its periodic adaptation and priority-based stealing. 
We isolate the impact of priority-based stealing by disabling it for \gml (results shown with the pink line in Figure~\ref{fig:nav-lat-default-wf}). %represents a setting where we disable priority-based stealing for the \gml for \oursys. 
Disabling priority-based stealing increases staleness, as the the scheduler cannot handle sudden spikes in \gml due to loop closure.

The above results show how \oursys performs better than the default configuration with respect to avoiding collisions and ensuring freshness of odometry readings. Improvements in these more critical metrics come at the cost of a small reduction in the area exploration rate -- the time taken to explore 90\% of the area increases by 10.9\%, as GM is allocated smaller amount of resources in \oursys than in the default scheme. Thus, \oursys achieves the desired trade-offs, as per the configured weights and priorities.

\begin{figure}[t]
\centering
%\setlength{\tabcolsep}{1pt}
%  \renewcommand{\arraystretch}{0.9}
%  \resizebox{\linewidth}{!}{
%   \begin{tabular}{cc}
    % \includegraphics[width=0.2\textwidth]{cs_thesis_latex_template-2020/figures/GM_run1.pdf}
    % &
     \includegraphics[width=0.37\textwidth]{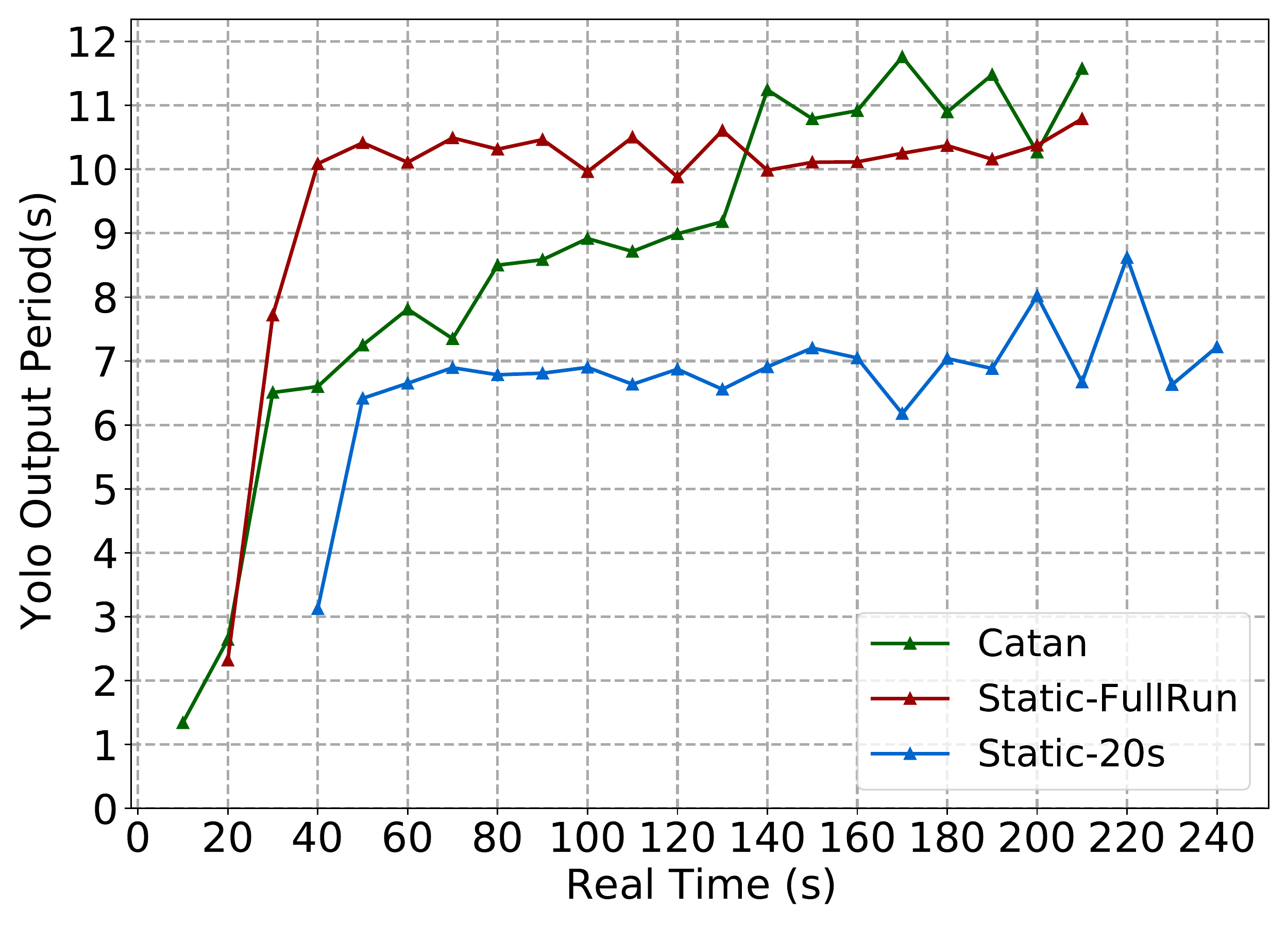}
     %&
     %\includegraphics[width=0.24\textwidth]{figures/Dyn_StaticFull_Static20s/Median_AvgLat_10sSlot_FullData.png}
    
 %   \end{tabular}}
 \vspace{-10pt}
    \caption{The output period (as a measure of reciprocal of throughput) for the YOLO chain across three schemes: \oursys, and two static baselines Static-FullRun and Static-20s on 1 core. We report the median of average period over 20 runs.}
\label{fig:nav2d-1c-yolo-stats}
\vspace{-10pt}
\end{figure}

\begin{figure}[t]
\centering
%\setlength{\tabcolsep}{1pt}
%  \renewcommand{\arraystretch}{0.9}
%  \resizebox{\linewidth}{!}{
%   \begin{tabular}{cc}
    % \includegraphics[width=0.2\textwidth]{cs_thesis_latex_template-2020/figures/GM_run1.pdf}
    % &
     \includegraphics[width=0.37\textwidth]{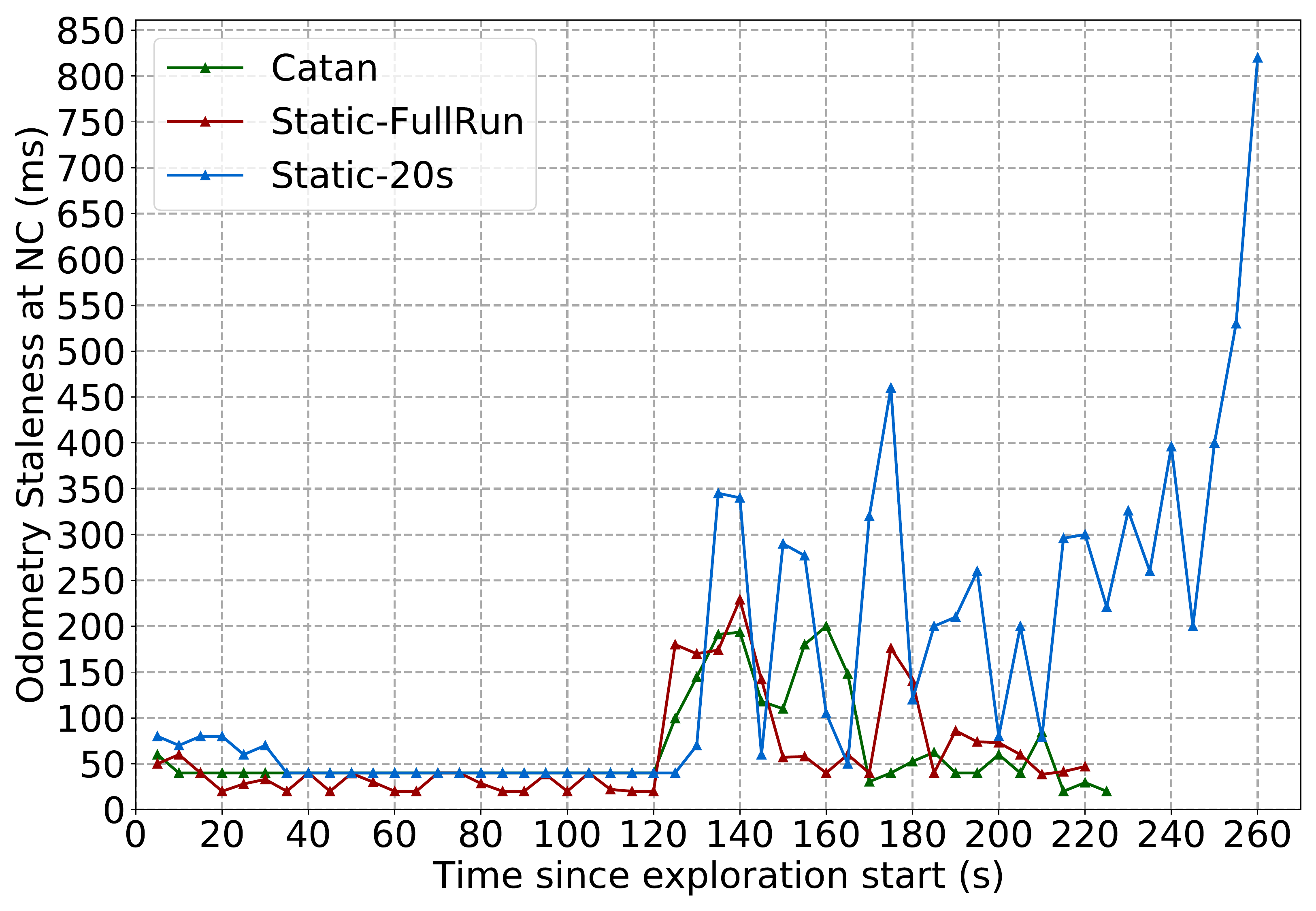}
    % &
    % \includegraphics[width=0.24\textwidth]{figures/Dyn_StaticFull_Static20s/XGTArea_MedRTime_Dyn_D20s_Stat.png}
    
%    \end{tabular}}
\vspace{-10pt}
    \caption{Comparing tail odometry staleness at NC (median over 20 runs) on Map2 across three schemes: \oursys, and two static baselines Static-FullRun and Static-20s on 1 core.}
\label{fig:yolo-1c-odom-area}
\vspace{-1em}
\end{figure}

\paragraphb{Adapting to variations in resource usage over time}
\label{sec:eval-nav2d-resrc-usage}
We next highlight the importance of updating scheduling decisions over time, as resource usage changes. 
We extend the navigation DAG by adding another chain that runs an object detector on camera images as the robot moves around (this is representative of robotic tasks, such as ObjectGoal task~\cite{anderson2018evaluation}). The chain comprises of a pre-processing node (similar to the one used in \S\ref{sec:facetracking-cs}) and a YOLO object detection node ~\cite{Redmon_yolo_2016_CVPR}\footnote{
While YOLO is usually GPU based, we
don't run it on GPU since we focus on CPU scheduling. We leave CPU/GPU
co-scheduling~\cite{ros-cpu-gpu-isorc} to future work.}. To model realistic timings in the simulated environment, we feed images from a real image dataset ~\cite{voc_dataset} to the object detection node.
% \aditi{While YOLO execution is usually GPU based, we run YOLO on the CPU for our evaluation. We leave CPU/GPU co-scheduling to future work.} 
We augment our objective function for \oursys to include the response time along this chain as another metric with a very low weight of 0.0005. 
We run this extended navigation application on a system with one core. 

\paragraphi{Baselines} We compare \oursys with two baselines that use a static configuration (that is not updated over time): (i) Static-FullRun that uses the same optimization problem as \oursys based on the tail computation time for each node over an entire experiment run on Map2 using a single core. It computes the schedule once, and does not re-solve it periodically. Note that since computation times of GM and GP nodes increase as the run progresses, this scheme uses a schedule derived from over-estimated computation time towards the beginning of the run. (ii) Static-20s is similar to Static-Full, but uses tail computation times of each node over the first 20s of the run. This scheme would use a schedule derived from an under-estimated computation time in the later half of the run. 

%Priority-based stealing for local chain nodes and \gml is enabled in both of the above baseline to handle sudden spikes in resource usage of higher priority nodes. For both baseline and for \oursys, YOLO is assigned the lowest priority (allowing local chain nodes and \gml to steal CPU resources from it if they overrun, in addition to stealing resources from other lower priority navigation nodes).  

\paragraphi{Metrics} We desire the objects to be detected in real time at a high rate, but that should not come at the cost of not being able to navigate well. Hence, we evaluate \oursys and the baselines on multiple metrics -- number of collisions in 40 runs on Map1 and, odometry staleness at \nc, area exploration rate and the average output period for the YOLO chain (as a measure of reciprocal of its throughput) over 20 runs on Map2. To aggregate YOLO chain's throughput data across 20 runs, we first take the average period (measured as the time gap between two consecutive outputs from the YOLO chain) over 10s buckets for each run, and then plot the median of these average values for each bucket.    

\paragraphi{Results}
Figure~\ref{fig:nav2d-1c-yolo-stats} shows the average output period for the YOLO chain over runs on Map2 -- lower is better. We find that Static-FullRun, which overestimates the computation time of navigation nodes in the first half of the run, assigns a low rate to the YOLO chain (resulting in high output period in the first half). Static-20s, which underestimates the computation of navigation nodes (by looking at the values in the first 20s of a run), assigns a high rate to YOLO -- as we discuss next, this comes at the cost of worse performance on a more critical metric. \oursys, with its dynamic re-solving assigns a high rate (low period) to YOLO at the beginning of the run, and then gradually decreases the rate (increases the period) over time. 

Figure~\ref{fig:yolo-1c-odom-area} shows the odometry staleness at navigation command across the three schemes run on Map2. While Static-FullRun and \oursys both perform similarly on this metric, Static-20s exhibits higher staleness towards the second half of the run. By using under-estimated computation times for the second half, Static-20s mis-allocates resources, giving a smaller share of CPU time to \gml, GM, and GP, and a more than necessary amount of CPU resources to the local chain nodes, NC, and YOLO. This reduces the effective throughput of \gml (in spite of priority-based stealing) and increases odometry staleness.

All three schemes performed similarly with respect to avoiding collisions on Map1, and in terms of the area exploration rate, and so we omit detailed results.

% \begin{figure}[t]
% \centering
% \setlength{\tabcolsep}{1pt}
%   \renewcommand{\arraystretch}{0.9}
%   \resizebox{\linewidth}{!}{
%   \begin{tabular}{cc}
%     % \includegraphics[width=0.2\textwidth]{cs_thesis_latex_template-2020/figures/GM_run1.pdf}
%     % &
%      \includegraphics[width=0.24\textwidth]{figures/Dyn_StaticFull_Static20s/NO_2cMedian of Yolo Output Period(per 10s Avg) vs time.pdf}
%      &
%      \includegraphics[width=0.24\textwidth]{figures/Dyn_StaticFull_Static20s/NO_2cMedian of Yolo Latency(per 10s Avg) vs time.pdf}
    
%     \end{tabular}}
%     \caption{Yolo Tput, Lat among dynamic and static-wholeRun and static-20s on 2core}
% \label{fig:nav2d-2c-yolo-stats}
% \vspace{-10pt}
% \end{figure}

\begin{figure}[t]
\centering
%\setlength{\tabcolsep}{1pt}
%  \renewcommand{\arraystretch}{0.9}
%  \resizebox{\linewidth}{!}{
%   \begin{tabular}{cc}
    % \includegraphics[width=0.2\textwidth]{cs_thesis_latex_template-2020/figures/GM_run1.pdf}
    % &
     \includegraphics[width=0.37\textwidth]{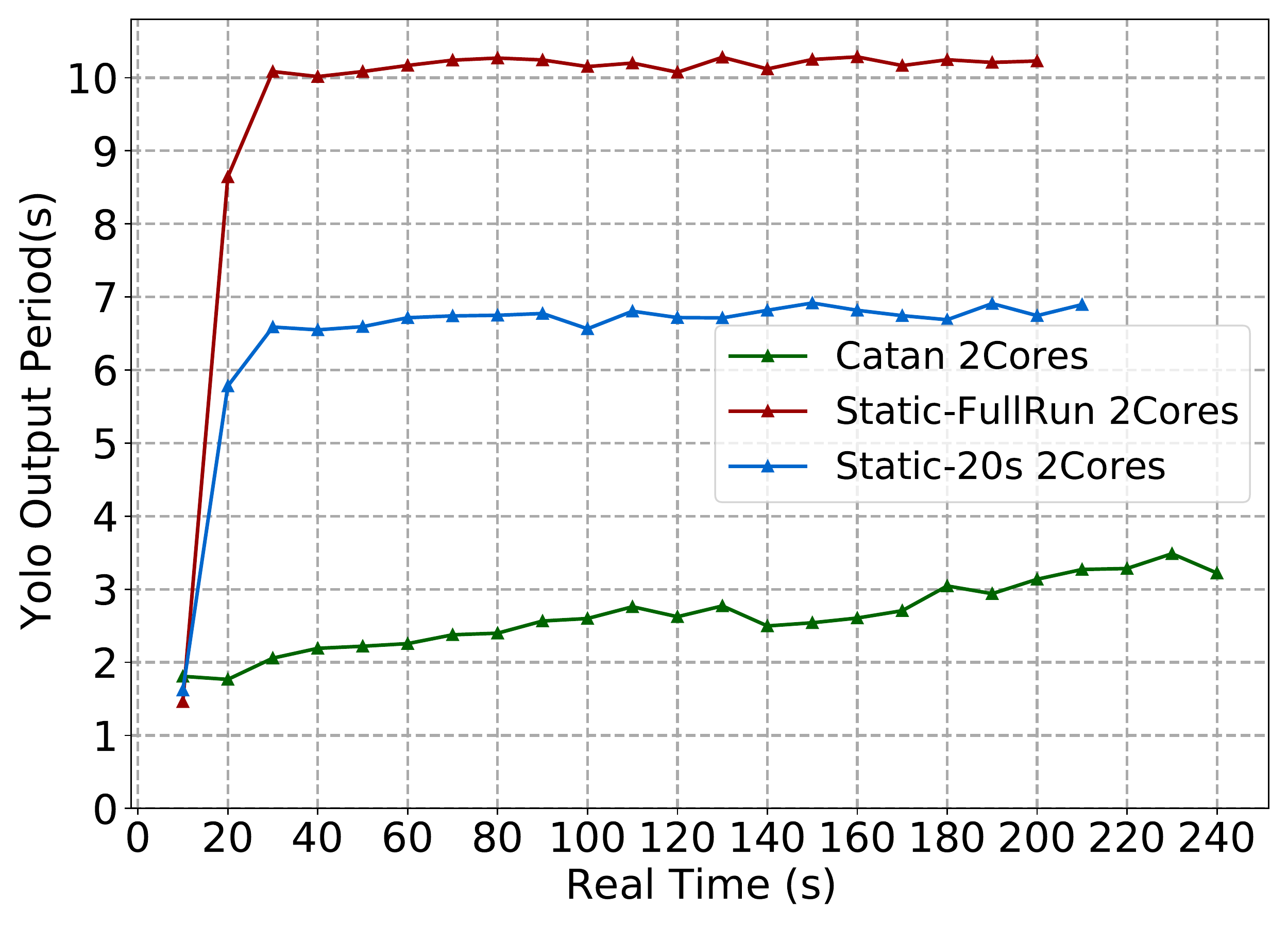}
 %    &
%     \includegraphics[width=0.24\textwidth]{figures/Dyn_StaticFull_Static20s/NO_1c_2cMedian of Yolo Latency(per 10s Avg) vs time.pdf}
    
%    \end{tabular}}
\vspace{-10pt}
    \caption{The output period (as a measure of reciprocal of throughput) for the YOLO chain across three schemes: \oursys, and two static baselines static-FullRun and Static-20s on 2 cores.}
\label{fig:nav2d-2c-yolo-stats}
\vspace{-10pt}
\end{figure}

\paragraphb{Adapting to change in resource availability}
Our experiments so far used a single core. We next ran the extended navigation application (with the YOLO chain) on Map2 on a system with two cores, to evaluate how well \oursys auto-scales. 

\paragraphi{Baselines} We configured the system to use the same node execution rates as those computed by the two baselines above (i.e. Static-FullRun and Static-20s with static schedules derived from node computation times on Map2 on one core), relying on default Linux policies to schedule the nodes across two cores.

\paragraphi{Metrics} We compare the average output period for the YOLO chain over time, as defined previously. 

\paragraphi{Results}
Figure~\ref{fig:nav2d-2c-yolo-stats} plots the average output period for the YOLO chain over time across \oursys and the two baselines on 2 cores. We find that \oursys is able to make efficient use of the extra core -- it dynamically allocates the core to just the local chain in the first half, and  the local chain and GP in the second half. It is, thus, able to run YOLO at a higher rate (and lower output period) than other static baselines that do not increase the execution rates. 
%Performance of \oursys and the two static baselines on two cores was similar with respect to the other metrics (and so we omit detailed results). 
% \aditi{slightly unfair to compare to baselines unaware of 2cores?}

\paragraphb{Key Takeaways} (i) Scheduling decisions impact how well a robot can navigate an environment while avoiding collisions. (ii) \oursys achieves the desired trade-offs across different metrics. (iii) It is important to adapt scheduling decisions over time and as resource availability changes -- \oursys is able to do so by dynamically re-solving the schedule over time. (iv) It is important to adapt to sudden fine-grained variability in resource usage -- \oursys is able to do so via priority based stealing.

% \begin{itemize}
%     \item Our scheme vs dynamic 1-shot vs static. 
%     \item Do we need to add a third chain for this experiment? 
%     1core ~\ref{fig:yolo-1c-odom-area} and ~\ref{fig:yolo-1c-tput-lat}.
%     1core collision map: 2runs/2collns with Dyn, 1run/1colln with Static, 1run/2collns with D20s.
    
%     2core collision map: 4run/5colln with Dyn, 2run/5colln with Static, 2run/4colln with D20s.
% \end{itemize}

\subsection{Virtual Reality}
\label{sec:vr-cs}
%Application overview

\paragraphb{Scheduler Configuration} We model the objective function as a weighted sum of response times for various chains in the VR DAG. We model IMU $\rightarrow$ Int as a subchain but do not explicitly schedule it since it has negligible compute time and executes at a high frequency. We configure \cam $\rightarrow$ VIO as a single subchain (since only VIO uses \cam 's output). The remaining nodes, \tw and \render are individual subchains. We use a weight of 1 and 0.5 for the chains IMU $\rightarrow$ Int $\rightarrow$ \tw and IMU $\rightarrow$ Int $\rightarrow$ \render $\rightarrow$ \tw respectively (these chains are more important, corresponding to rotational and translational motion-to-photon response time respectively). For all the remaining (4) chains that go through VIO, we use a weight of 0.005. We also add throughput constraints as per the display refresh rate (vsync, determined by the display hardware and driver): (i) \tw's rate is upper and lower-bounded by vsync.
% \sva{so does this not fix it to be vsync?} \aditi{it does, but we represent constraints as < and > in our formulation.}
(ii) \render's rate is upper-bounded by vsync.
% \sva{why not lower bounded by vsync too like timewarp?} \aditi{since if there are limited resources, we might want to slow it down}
%, since there is no benefit to executing Render faster than the driver's display rate. 
The order of the subchains is specified to be \tw followed by \cam $\rightarrow$ VIO, then \render, so that each render output is as close to the next timewarp execution as possible. We enable priority-based stealing only for timewarp, 
%\footnote{We don't enable stealing for render since there is no benefit to running render in the  \cam $\rightarrow$ VIO subchain's cpu timeslice which is just after timewarp (vsync) in the configured order}, 
i.e. it is not preempted and can steal CPU time from the other nodes if needed.
% \radhika{replace C, TW and R with Camera, Timewarp and Render above}

\paragraphb{Experiment Setup}
We use the ILLIXR system for our VR application~\cite{HuzaifaDesai2021,illixr-site}. 
% to run the VR application and generate the graphics which would be shown on a headmounted display. 
% To improve reproducibility, we drive the system using a dataset instead of real sensors. 
As discussed in \S~\ref{sec:vr-bg}, we use a standard dataset for the camera and IMU inputs to ILLIXR for repeatability. We use Vicon Room 1 Medium from EuRoC~\cite{BurriNikolic2016}, which consists of an 85 second long trajectory, and provides camera images at 20 Hz and IMU readings at 200 Hz.
% We add an artificial node to the system that models the sensors by reading from the dataset.
% Using a fixed dataset input improves reproducibility, but it implies adding an "artificial" node that models the sensors. We omit the impact that this node has on the chain latencies, since it would not be present in an actual system.
The Render component renders a custom scene consisting of a living room with furniture and textured walls. We use a display with a resolution of 2560x1440, and a refresh rate of 120 Hz (vsync period of 8.33 ms). We run the CPU parts of our setup on a single core of an Intel i9-10900K CPU clocked at 5.3 GHz. The GPU parts execute on an NVIDIA GeForce RTX 3090 with a core clock of 2.1 GHz.
% \aditi{We repeat the runs ? times for all the 3 settings, and present aggregate results}

\paragraphb{Baselines} We compare \oursys against two baselines. The first baseline configures static rates (\tw and \render running at vsync frequency and \cam running at the frequency of the image dataset) and relies on Linux default policy SCHED\_OTHER for all other scheduling decisions. The second baseline extends the first one, by using SCHED\_FIFO for all the nodes and prioritizing the execution of different components in the following order: IMU and Int (so that they run at the dataset's high frequency),
% \sva{why is the requirement of running at dataset frequency related to first in the order?}) \aditi{similar to how we assign imu,int high priority even in the scheduled case, so it can preempt all other nodes when needed}
followed by TW, then Render and then Camera and VIO (based on decreasing order of their chains' weights). 
% \sva{Is there a better justification for this order?} \aditi{no - we expect weights' ordering to reflect semantic importance order, so just using the weights}
% The node emulating the IMU sensor is always given a high priority, to ensure that accurately emulates a physical sensor.

\paragraphb{Metrics} We compare the response time along the chains (A) IMU $\rightarrow$ Int $\rightarrow$ \tw (that captures rotational MTP), (B) IMU $\rightarrow$ Int $\rightarrow$ \render $\rightarrow$ \tw (that captures translational MTP), and (C) \cam $\rightarrow$ VIO $\rightarrow$ Int $\rightarrow$ \tw  (that captures how quickly accurate pose-information is available from VIO). For each scheme, we first average the metrics within each run over time, and then depict the average and standard deviation of these values across 20 runs.

\begin{figure}
    \centering
    \includegraphics[width=0.35\textwidth]{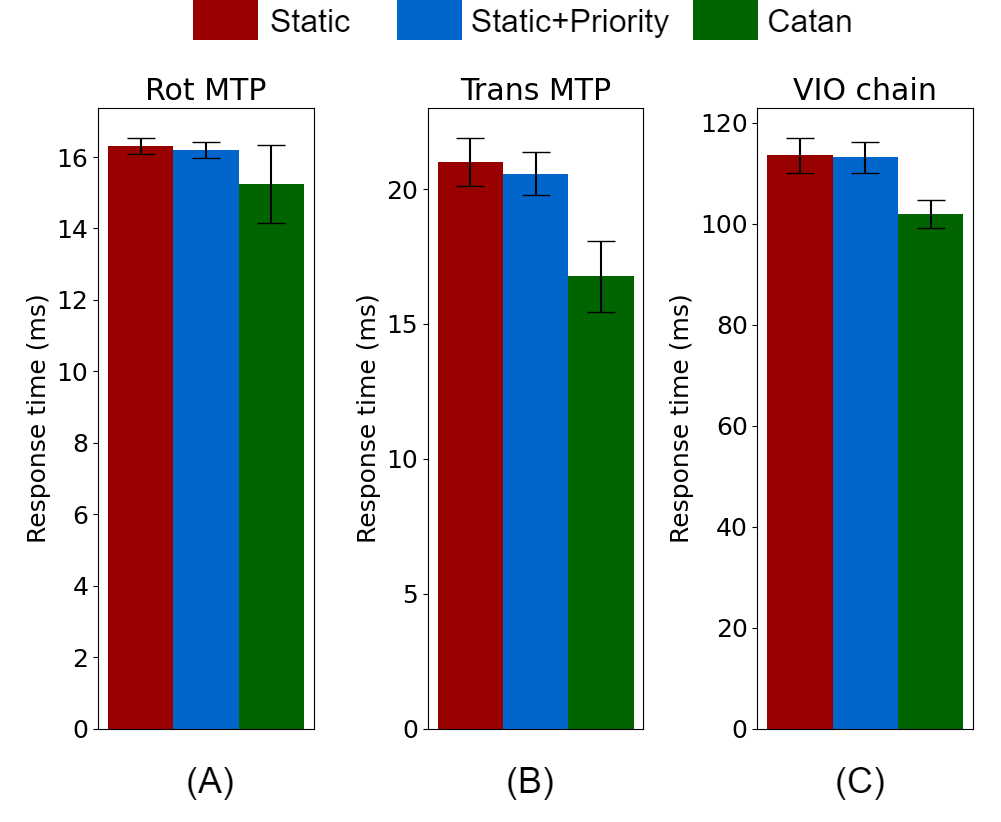}
    \vspace{-10pt}
    \caption{ Response time of the chains processing (A) approx. rotational motion, (B) approx. translational motion, and (C) accurate motion via VIO. The error bars show one standard deviation above and below. These results are aggregated over 20 runs.}
    \vspace{-10pt}
    \label{fig:illixr-all-rts}
% \vspace{-2em}
\end{figure}

\paragraphb{Results}
Figure \ref{fig:illixr-all-rts} compares the two baselines with Catan.
%Chain A is the path which determines the response-time for capturing rotational motion, B determines response-time for translational motion, and C determines the response-time for accurate pose-information from VIO.
% , as it goes through the costly but accurate VIO component.
%e choose these three chains to highlight how well the system tracks the rotational and translational motion and
% There is significant compute time variation in the components being scheduled. The dynamic scheduler is able to learn this variation. 
% As a result, it eventually converges on a scheduling policy which improves the core metrics, rotational MTP (\cref{fig:rot-mtp}), translational MTP (\cref{fig:trans-mtp}), and VIO response time (\cref{fig:illixr-vio-rt}). 
%To make our analysis robust to variance between runs of the same configuration, we aggregate over mulitple runs.
%~\footnote{The intra-run variance over time (not reported for brevity) presents to the user as minor hiccups -- it is about 1.5$\times$ higher for Catan for chains A and B than the baselines. \radhika{this might be misleading and make us look unecessarily bad -- what does 1.5$\times$ higher mean on an absolute scale?} We focus on the aggregated response times over a run, a higher value for which presents to the user as a \emph{persistently} bad experience for that run. } 
%Therefore, we consider inter-run variance, reflected by the standard deviation of averages, of greater significance.}
We make the following observations: (i) The priority stealing mechanism in \oursys enables it to execute \tw just before vsync in-spite of variations in computation times of different nodes, leading to better \rt by \(\approx\)7\%. (ii) \oursys's temporal CPU allocation to \render ensures that it finishes execution as close to vsync as possible, leading to \(\approx\)25\% improvement in chain B \rt. (iii) Lastly, our system is able to trigger \cam at a high rate,\footnote{Even though the offline dataset has a fixed camera input rate, sending triggers for camera at a higher rate reduces delay due to asynchrony between offline camera inputs and triggers. We expect to see similar (and potentially larger) benefits with live camera inputs.} while continually adapting to varying compute times, without affecting the rest of the system due to its explicitly planned schedule, leading to \(\approx\)11\% improvement in chain C.
% \sva{How can the camera run at a higher rate than provided by the image dataset? Earlier it was mentioned that the baseline is runnign at the dataset rate.}\aditi{added footnote}
% triggers Cam at high rate, while not affecting [explicit scheduling helps?].
% \aditi{ability to allocate cpu without affecting the tw/render}
Overall, \oursys is able to make the desired trade-offs across different components in the application.
% \sva{Are the above explanations clear by this time in the paper or do we need some more handholding for the reader to understand the results?}

% \begin{figure}[t]
% \centering
%     \includegraphics[width=0.45\textwidth]{figures/rotational_mtp.png}
%     \caption{The latency between rotational-motion occurring in the real world, and that motion being displayed in the virtual world.}
%     \label{fig:illixr-rotational-mtp}
%     \vspace{-10pt}
% \end{figure}

% \begin{figure}[t]
% \centering
%     \includegraphics[width=0.45\textwidth]{figures/rotational_mtp.png}
%     \caption{The latency between translational-motion occurring in the real world, and that motion being displayed in the virtual world.}
%     \label{fig:illixr-translational-mtp}
%     \vspace{-10pt}
% \end{figure}

% \begin{figure}[!t]
%     \subfloat[Rotational]{
%     \includegraphics[width=\linewidth]{figures/rotational_mtp.png}
%     \label{fig:rot-mtp}
%     }
%     \vspace{-10pt}
%     \subfloat[Translational]{
%     \includegraphics[width=\linewidth]{figures/translational_mtp.png}
%     \label{fig:trans-mtp}
%     }
%     % \captionsetup{font=small}
%     \caption{
%     The latency between rotational (a) and translational (b) motion occuring in the real world, and the motion being displayed in the virtual world. \aditi{is this avg latency? over 2 runs?}
%     }
%     \label{fig:mtp}
% \end{figure}

% \begin{figure}[t]
% \centering
%     \includegraphics[width=0.45\textwidth]{figures/vio_rt.png}
%     \caption{The response time of VIO.}
%     \label{fig:illixr-vio-rt}
%     \vspace{-10pt}
% \end{figure}
\section{Related Work}
\label{sec:related}

System scheduling has a rich literature. We discuss three broad (and most closely related) categories, highlighting their limitations in addressing the requirements for \sr systems. 

\paragraphi{Real-time scheduling} Prior work in real-time scheduling for DAGs assumes static periodic schedules, and focus on meeting latency constraints under given (and fixed) time periods at which a node is triggered~\cite{openvx-realtime, realtime-parallel-sched, dag-scheduling-rtc, yang-priorities} (including the few that look at robotics applications \cite{rosch-rt-e2e-lat, verucchi2020latency, picas-rtas-21, rt-exec-ros2-icess}). Davare et al.~\cite{davare2007period} focus on assigning per-node time periods in automotive applications 
% \fixme{application domain?} 
but under given node priorities and core allocations. These works use expensive algorithms, and do not account for variability in processing times.
%\fixme{add cite. if all fall in this category, edit sentence to say all} \fixme{Further comment from Brighten: if not all fall into the category, maybe split them out and discuss them separately.  I agree saying ``most'' is not enough.}.
%~\cite{rosch-rt-e2e-lat, realtime-dag-model, realtime-parallel-sched, dag-scheduling-rtc, openvx-realtime, davare2007period}. 
%\radhika{Aditi, check if this is correct, especially the last sentence. Also, add any other cites we might be missing here} 

\paragraphi{Stream-processing systems} Many stream processing engines make scheduling decisions locally at each node (which are often suboptimal)~\cite{seda, borealis, heron, zeromq, rabbitmq, seda-followup, streamscope}. Among those that make system-wide decisions, most focus on resource allocation or task scheduling to handle a given (though variable) input load (e.g.~\cite{aurora-scheduling, sched-variable-load, dfg-naiad, dfg-dryad}). %\aditi{ok to dataflow cites here?}. 
A few look at input load shedding to avoid system overload, but do not simultaneously optimize resource allocation and task ordering (e.g.~\cite{aurora-load-control, load-shedding-data-streams}). More generally, such mechanisms are designed for a different application domain and environment (e.g. query processing in cloud clusters), and do not optimize for the application-level goals in resource-constrained \sr systems. 
%\fixme{Any other requirements they fail to meet?}

\paragraphi{Sensor Nodes} Pixie ~\cite{sensornode-pixie} is an operating system for sensor nodes, which provides a dataflow programming model with abstractions for managing resources - but leaves all the scheduling decisions upto the application. Flask ~\cite{sensornode-flask} also uses a dataflow model to define a Domain Specific Language for sensor network applications, but does not handle variations in compute usage or availability. While Eon ~\cite{sensornode-eon} provides a flexible interface to express how the scheduling parameters can be adapted at runtime, it requires the developer to specify how to change each parameter based on different resource availability. More generally, none of the sensor node systems deal with complex compute or fine grained variability.

% \aditi{Todo: add DFG [Naiad, Dryad, RaftLib], Sensor nodes [pixie,flask,eon]}
% \paragraphi{Sensor Networks} 
% Similar resource management problems in networks of sensor nodes, 
\paragraphi{Offline profiling} Some of the data-driven approaches rely on extensive offline profiling to tune different aspects of the system configuration (e.g.~\cite{automagic, videostorm, opentuner, auto-clustersz-hadoop, bo4co}). However, such approaches cannot adapt to dynamic variations over time and across deployment scenarios. Nonetheless, extending such data-driven approaches to handle the requirements of \sr systems is an interesting future direction.
%\fixme{other cites?}
\section{Conclusion}
\label{sec:future}

%\radhika{rename section to Discussion / Future Work}

In this work, we build an understanding of the scheduling requirements and challenges for \sr systems by studying three applications spanning robotics and VR. Using this understanding, we develop a scheduler to manage on-device CPU resources for \sr applications. We show how it is able to handle multiple scheduling dimensions across heterogeneous app nodes, adapt to variations in compute resource usage at different timescales, and achieve the specified semantic trade-offs.
%space as per the specified configuration

We believe our work to be a first step in a rich problem domain of \sr system scheduling, with multiple interesting questions open for future research.
%, such as studying more  
%For instance, \oursys currently requires substantial inputs from app developers -- the weighted objective function and constraints on low-level system metrics, the granularity at which scheduling decisions are made for the app components, and which nodes are allowed to pre-empt which other nodes. An interesting future direction is to infer these inputs to further ease the task of app developers. For instance, the mapping between app-level performance goals and low-level system metrics can be potentially learnt through data-driven approaches. We expect such a mapping to generalize better across scenarios and over time than directly learning scheduling configurations.  
% \aditi{Added future directions:}
For instance, \oursys currently requires several inputs from the app developers, e.g. performance constraints and objectives along different components of an application. While we believe that specifying such high-level inputs is still significantly easier than configuring low-level system knobs (as required today), an important direction for future research includes automatically inferring these inputs (e.g. via data-driven approaches) to further ease app development.   

As mentioned in \S\ref{sec:background}, extending our scheduler (currently restricted to managing CPU resources on a single platform) to handle other resources (e.g. GPU, accelerators, etc), varying deployment models (e.g. tasks offloaded to edge or cloud servers) as well as different frameworks (e.g. ROS 2) is another interesting future direction.  

%\radhika{extend this further. }
\section*{Acknowledgements}
This work was supported by Intel, AG NIFA under grant 2021-67021-34418, the Applications Driving Architectures (ADA) Center, a JUMP Center co-sponsored by SRC and DARPA, the DARPA DSSOC program, and NSF under grants 2120464 and 2008971.

%\newpage

%\radhika{old text begins from here}

%\section{Introduction}
%\input{old/introduction-old}

%\section{Background}
%\input{old/background-old}

%\section{Unique Challenges?}

%\section{Related Work?}

%\section{\oursys Design}
%\input{old/design-old}

% \section{Implementation}
% \input{implementation}

% \section{Evaluation}

% \section{Conclusion \& Future Work}

\bibliographystyle{plain}
\bibliography{refs}

\end{document}